\documentclass[11pt]{article}

\usepackage[preprint]{acl}
\usepackage{times}
\usepackage{latexsym}
\usepackage[T1]{fontenc}
\usepackage[utf8]{inputenc}
\usepackage{microtype}
\usepackage{inconsolata}
\usepackage{graphicx}
\usepackage{amsfonts}   
\usepackage{amsmath}
\usepackage{algorithm}
\usepackage{algorithmic}
\usepackage{booktabs}
\usepackage[table]{xcolor}
\definecolor{authorcolor}{RGB}{0,0,0}
\usepackage{multirow}
\usepackage{array}
\usepackage{siunitx}
\usepackage{enumitem}
\usepackage{marvosym}
\newcolumntype{Y}{>{\centering\arraybackslash}p{0.62cm}}
\newcolumntype{M}{>{\centering\arraybackslash}m{2.8cm}}
\newcolumntype{A}{>{\centering\arraybackslash}m{0.95cm}}
\newcolumntype{L}{>{\raggedright\arraybackslash}m{2.8cm}}

\title{Stability Implies Redundancy: Delta Attention Selective Halting \\ for Efficient Long-Context Prefilling}

\author{Yujie Chen$^{1}$, Tailai Chen$^1$, Yifeng Gao$^{1}$, \\
\textbf{Zoe Wanying He$^{2}$, Yijue Xu$^{3}$, Shaobo Wang$^{1}$, Linfeng Zhang$^{\text{\Letter}}$$^{1}$} \\
$^{\color{authorcolor}\boldsymbol{1}}$ Shanghai Jiao Tong University $\quad$ \\
$^{\color{authorcolor}\boldsymbol{2}}$ University of California, San Diego $\quad$ 
$^{\color{authorcolor}\boldsymbol{3}}$ Carnegie Mellon University $\quad$ \\
{
\Letter\ Corresponding author  $\quad$
}}

\begin{document}

\maketitle
\begin{abstract}
Prefilling computational costs pose a significant bottleneck for Large Language Models (LLMs) and Large Multimodal Models (LMMs) in long-context settings. While token pruning reduces sequence length, prior methods rely on heuristics that break compatibility with hardware-efficient kernels like FlashAttention. In this work, we observe that tokens evolve toward \textit{semantic fixing points}, making further processing redundant. To this end, we introduce Delta Attention Selective Halting (DASH), a training-free policy that monitors the layer-wise update dynamics of the self-attention mechanism to selectively halt stabilized tokens. Extensive evaluation confirms that DASH generalizes across language and vision benchmarks, delivering significant prefill speedups while preserving model accuracy and hardware efficiency. Code will be released at \url{https://github.com/verach3n/DASH.git}.
\end{abstract}
\section{Introduction}
The capability to process extensive context windows is a key feature of Large Language Models (LLMs) and Large Multimodal Models (LMMs), enabling long-document summarization and high-resolution visual understanding \citep{gpt4,geminifamily}. However, this capability comes with a prohibitive computational cost. While techniques such as quantization \citep{llmint8} and FlashAttention \citep{flashattention} have alleviated memory bottlenecks, the prefill complexity remains quadratic or linear with respect to sequence length. As token sequences grow into the hundreds of thousands, the floating-point operations required for the forward pass become the primary latency bottleneck.

To address this problem, recent research has explored token pruning strategies to reduce the sequence length \citep{h2o}. However, most existing methods rely on heuristic importance scores (e.g., accumulated attention weights) that require access to the attention matrix, making them incompatible with FlashAttention and other efficient attention operators. This leads to a question: instead of asking \emph{which tokens are important}, can we identify \emph{which tokens have already finished their job}?

To answer this question, we revisit token dynamics across layers through the lens of residual connections, where $x_{l+1}=x_l+\Delta$. 
We posit that tokens evolve towards a ``semantic fixed point'', where a diminishing update magnitude $\Delta$ signals that the representation has stabilized within the current context. Specifically, we monitor the update of the self-attention layer ($\Delta_{attn}$ ) to capture global information exchange, rather than local feature variations.

\begin{figure}[!t]
  \centering
  \includegraphics[width=0.9\linewidth]{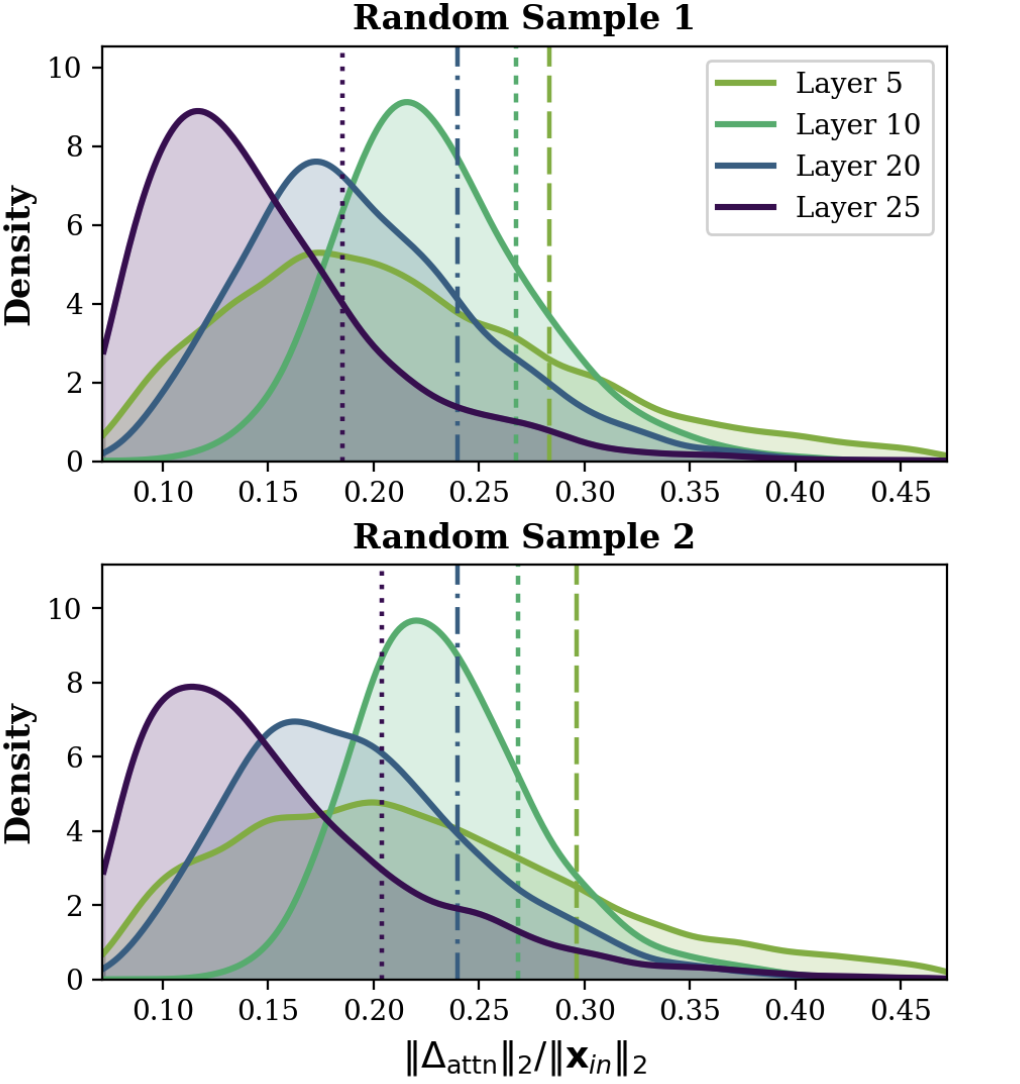}
  \caption{Layer-wise distributions of token-wise relative $\Delta_{\mathrm{attn}}$ at layers 5, 10, 20, and 25 for Qwen2.5-7B-Instruct-1M, computed on two randomly sampled LongBench-E (2WikiMQA) instances.}
  \vspace{-0.5cm}
  \label{fig:intro_delta_hist}
\end{figure}

\noindent\textbf{Observation 1: Semantic fixing points exist.} 
This hypothesis is validated by the empirical distribution of $\Delta_{attn}$, as
  shown in Figure~\ref{fig:intro_delta_hist}. We observe a highly skewed pattern where the vast majority of tokens cluster near zero, indicating they have effectively reached their semantic fixed points. Only a sparse subset of tokens continues to undergo significant updates in deeper layers. This distinct separation reveals a substantial computational redundancy: once a token enters this ``saturation regime,'' further processing contributes little to the global context and yields diminishing returns.

\begin{figure}[!t]
  \centering
  \includegraphics[width=\linewidth]{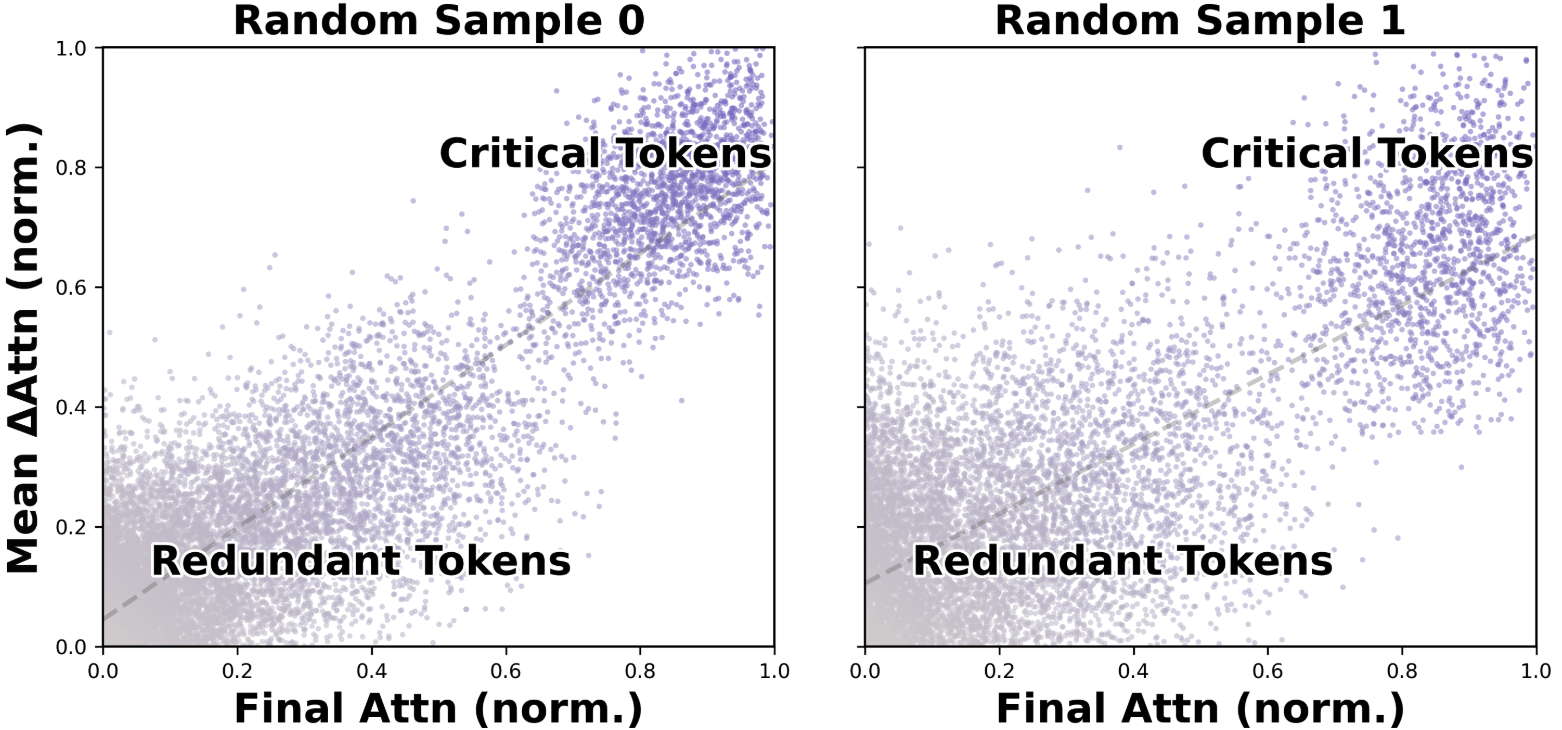}
  \caption{Correlation between normalized final-layer attention scores (x-axis) and the mean Delta Attention \(\Delta_{\text{attn}}\) of each token across layers (y-axis).}
  \vspace{-0.5cm}\label{fig:intro_corr}
\end{figure}

\noindent\textbf{Observation 2: Tokens converged to their fixed point are redundant.} To confirm that the semantic fixed point'' identified in Observation 1 translates to a loss of influence over the global context, we analyze the correlation between update magnitude and token importance. Figure~\ref{fig:intro_corr} plots each token's normalized attention score against its \(\Delta_{\text{attn}}\), revealing a strong positive relationship: tokens that stabilize (low \(\Delta_{\text{attn}}\)) rarely attract significant attention from subsequent layers. This behavior suggests that once a token saturates, it effectively becomes an information sink,'' allowing us to use 
$\Delta_{attn}$
  as a safe, training-free proxy for pruning without computing expensive attention matrices.


\begin{figure}[!t]
  \centering
  \includegraphics[width=\linewidth]{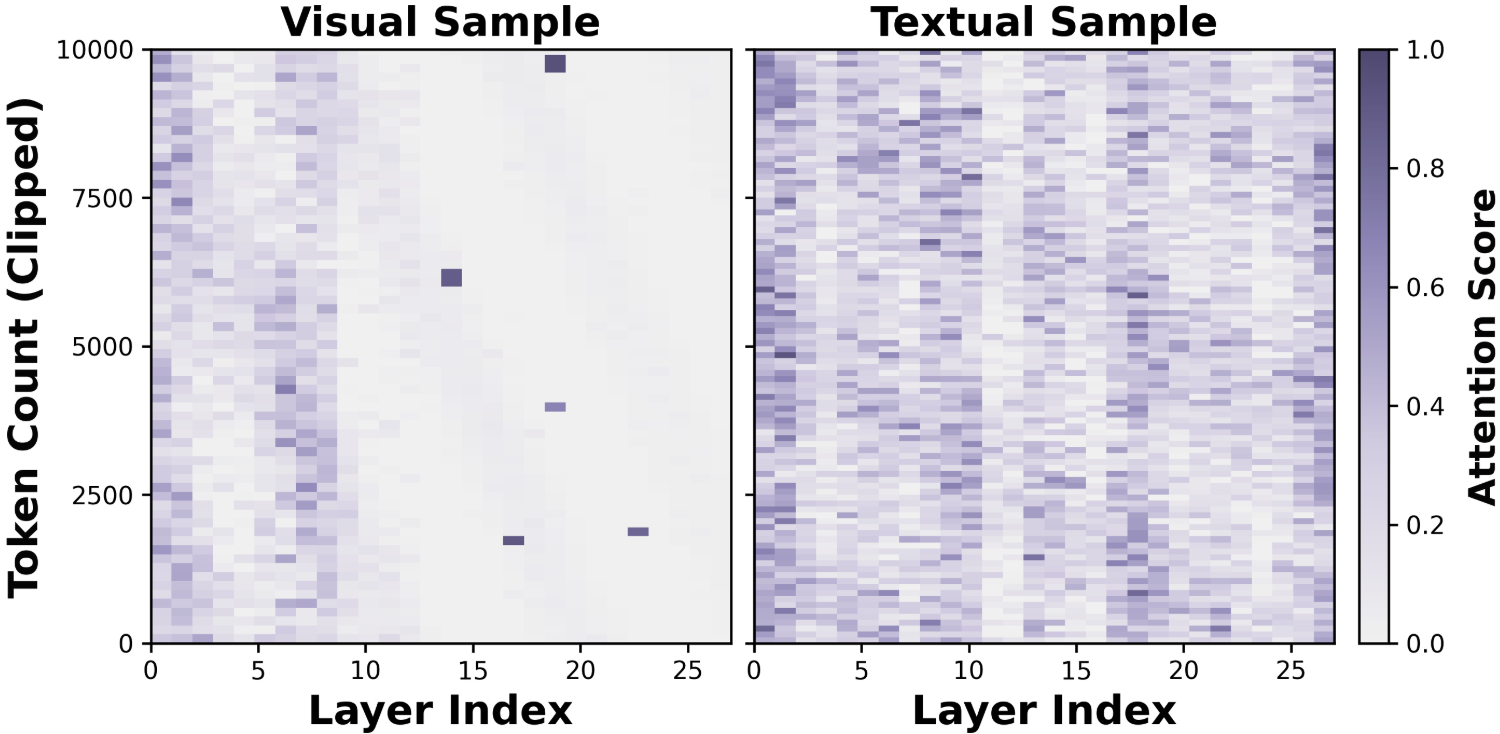}
  \vspace{-0.7cm}
  \caption{{Layer-wise sparsity patterns} for visual (left) and textual (right) tokens, showing earlier saturation for visual tokens compared with text tokens.}
  \label{fig:intro_sparsity}
  \vspace{-0.5cm}
\end{figure}

\noindent\textbf{Observation 3: Visual tokens saturate earlier than textual tokens.} While saturation serves as a general indicator of redundancy, the \textit{rate} of convergence differs fundamentally across modalities. Figure~\ref{fig:intro_sparsity} compares the layer-wise attention sparsity patterns for visual and textual tokens within a vision-language model. We observe a distinct disparity: visual tokens tend to reach their semantic fixed points and exhibit high sparsity significantly earlier in the network, whereas textual tokens require deeper layers to process and stabilize~\\citep{fastv}.
This modality gap explains a common pitfall in efficient inference: token pruning strategies originally developed for large multimodal language models
often fail when directly applied to Large Language Models, calling for token pruning methods designed for both language and vision.


Building on these insights, we propose \textbf{DASH} (Delta Attention Selective Halting),  a training-free inference-time policy for fast prefill. DASH computes token-wise $\Delta_{\text{attn}}$ once at a start layer $l_s$, selects a compact active set by retaining the $top K=\lfloor(1-\rho)T\rfloor$ tokens, and reuses this fixed set for all subsequent prefill layers. Tokens outside the active set are halted at $l_s$ and skip both self-attention and FFN computation in deeper layers. This single-shot schedule is compatible with efficient attention operators (e.g., FlashAttention \citep{flashattention}) since it does not require materializing full attention matrices.
We evaluate \textsc{DASH} on both long-context text and vision--language benchmarks. Across these settings, \textsc{DASH} consistently outperforms representative prefill-time token compression baselines under the same token reduction setting, while remaining close to the uncompressed backbone. Moreover, the vision--language results exhibit a larger performance margin under more aggressive reduction ratios, indicating that \textsc{DASH} retains accuracy more effectively when prompts are inflated with dense visual tokens.

In summary, our contributions are three-fold:
\begin{itemize}[leftmargin=10pt, topsep=0pt, itemsep=1pt, partopsep=1pt, parsep=1pt]
    \item We identify per-layer \(\Delta_{\text{attn}}\) as a first-order indicator of token saturation and provide empirical evidence of its correlation with token importance and the difference between vision and language (Figures~\ref{fig:intro_delta_hist}, ~\ref{fig:intro_corr}, and Figure~\ref{fig:intro_sparsity}).
    \item We introduce \textbf{DASH}, a training-free framework that leverages \(\Delta_{\text{attn}}\) to perform single-shot selective token halting at $l_s$, substantially reducing prefill latency without retraining.
    \item Extensive experiments on long-context language and vision--language benchmarks with Qwen2.5-7B-Instruct-1M and Qwen2-VL-7B validate \textbf{DASH}, delivering consistent prefill efficiency gains with minimal performance degradation and strong accuracy--efficiency trade-offs against competitive baselines.
\end{itemize}

\section{Related Work}
\subsection{Memory-Oriented Inference} Memory-oriented approaches primarily aim to compress the key-value (KV) cache to reduce memory footprint and bandwidth. Common strategies include eviction policies based on accumulated attention or recency \citep{h2o, snapkv, keyformer}, and more refined criteria leveraging head-level or value-aware signals \citep{fastgen, vatp}. While effective for decoding efficiency, these methods offer limited savings for the dominant prefill computation, as the full prompt must typically be processed to construct the initial cache. Recent work further replaces static heuristics with graph-based dynamic importance propagation for KV cache eviction \citep{graphkv}.

\subsection{Structural Compute Reduction} Complementary to cache compression, structural approaches reduce inference FLOPs by pruning the computation graph. Techniques range from exploiting specific attention patterns like attention sinks \citep{attentionsink} to dynamic token pruning strategies \citep{lazyllm, sliminfer, adtp} and depth-wise early exiting \citep{dtop}. While token pruning is well-explored in vision \citep{tps}, recent work has also extended training-free token reduction to multimodal settings \citep{ficoco} and  VideoLLM acceleration \citep{vidcom2, stc}. However, training-free token reduction for autoregressive text generation remains understudied. Our work addresses this gap by proposing a lightweight policy that targets prefill FLOPs and generalizes to multimodal inputs.

\subsection{Information Propagation and Redundancy} Recent analyses of deep Transformers reveal significant redundancy, where deeper layers often introduce only minor representation updates \citep{layerredundancy}. Mechanistic studies further suggest that attention heads perform iterative computation, gradually refining intermediate states \citep{attentionflow, emergentroles}. These findings motivate shifting from static importance metrics to criteria based on dynamic updates. Accordingly, we leverage the magnitude of the attention residual update as a layer-local proxy for marginal computational return to guide our training-free selective halting. Beyond autoregressive models, similar observations of token stability have been exploited to accelerate diffusion-based language models via adaptive caching \citep{dllmcache}.

\section{Methodology}
\begin{figure*}[h]
  \includegraphics[width=\linewidth]{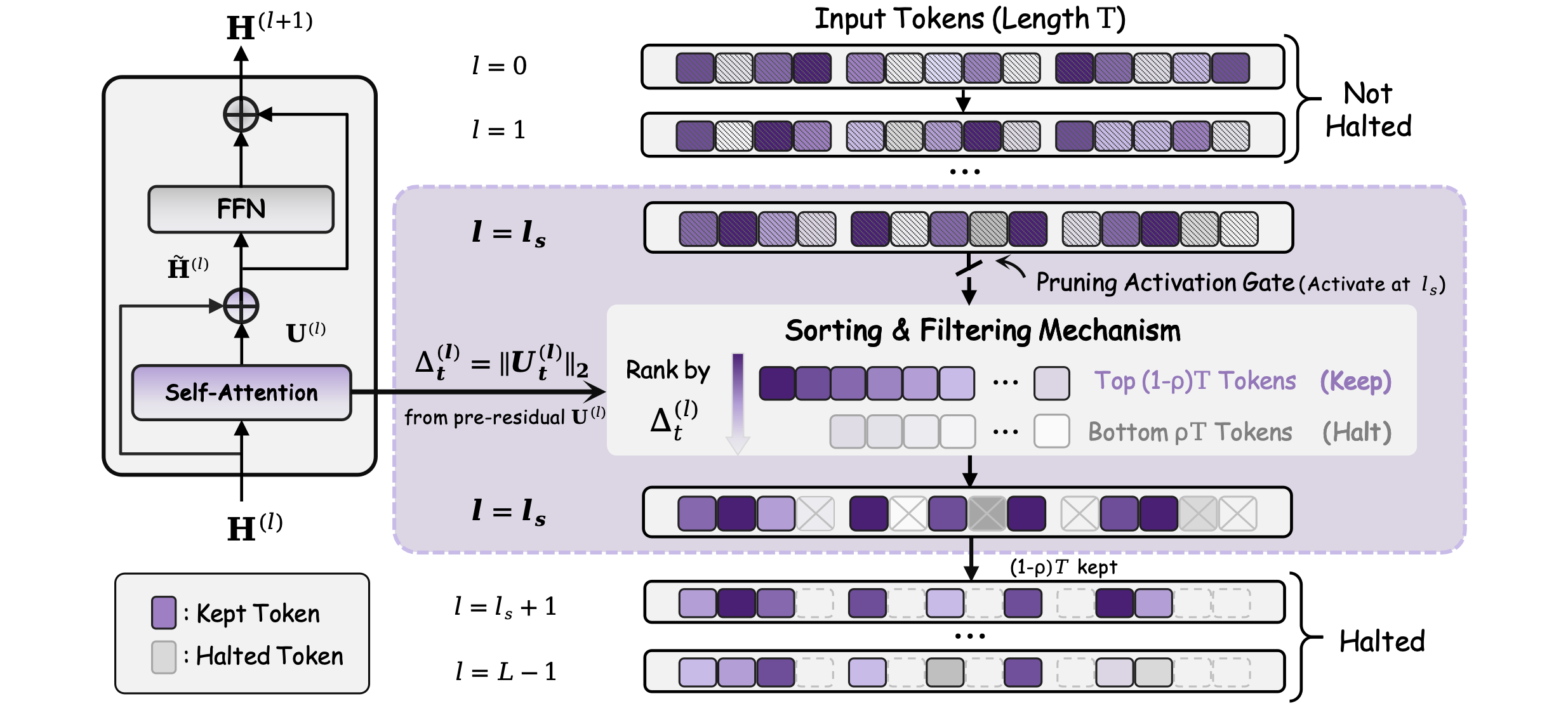} 
  \caption{
    Overview of \(\Delta_{\text{attn}}\) guided token halting during prefill.
    \textbf{Left:} in block $l$, self-attention produces the pre-residual output $\mathbf{U}^{(l)}$, and we define a per-token \(\Delta_{\text{attn}}\) score $\Delta_t^{(l)}=\lVert \mathbf{U}^{(l)}_t\rVert_2$.
    \textbf{Right:} given an input of length $T$, all tokens are processed normally for layers $l<l_s$.
    At the activation layer $l_s$, we rank tokens by $\Delta_t^{(l_s)}$ and apply a fixed pruning ratio $\rho$, keeping the top $(1-\rho)T$ tokens (purple) and halting the remaining tokens (gray).
    Halted tokens are removed from the active set and, in all deeper layers, \emph{skip the entire Transformer block computation}, including both self-attention and the feed-forward network (FFN), thereby reducing prefill FLOPs.
    The figure highlights the kept tokens (filled purple), halted tokens (gray with cross), and the pruning gate activated at $l_s$.
    }
    \label{fig:overview}
\end{figure*}

\subsection{Preliminaries}
\paragraph{\textbf{Long-context and multimodal inference.}}
Inference for autoregressive Transformers consists of a \emph{prefill} stage and a \emph{decoding} stage. Given an input sequence of length $T$, prefill processes all $T$ tokens once to produce hidden states and initialize the KV cache, after which decoding generates tokens autoregressively while reusing cached keys and values. In long-context settings, prefill can dominate latency because it executes attention and feed-forward computation over the entire prompt, and the cost grows rapidly with $T$ (quadratically for self-attention). In practice, $T$ is often inflated beyond plain text due to retrieval-augmented prompts and multimodal inputs with dense visual tokens, further increasing time-to-first-token. We therefore focus on reducing prefill computation for both long-context text generation and multimodal generation.

\paragraph{\textbf{Transformer blocks.}}
We consider a Transformer with $L$ blocks. Let $\mathbf{H}^{(l)} \in \mathbb{R}^{T \times d}$ denote the token representations entering the self-attention sublayer at block $l$. The self-attention output \emph{before} residual addition is denoted by $\mathbf{U}^{(l)} \in \mathbb{R}^{T \times d}$, yielding the residual update $\widetilde{\mathbf{H}}^{(l)}=\mathbf{H}^{(l)}+\mathbf{U}^{(l)}$.

\paragraph{\textbf{Token pruning and pruning ratio.}}
Token pruning implements conditional computation by halting a subset of tokens in intermediate and deep layers, so that only the remaining tokens are propagated through subsequent blocks. We denote by $\rho \in [0,1)$ the \emph{pruning ratio}, meaning that $(1-\rho)T$ tokens are retained at layers where pruning is activated. Prior evidence suggests that information can diffuse across depth, motivating depth dependent selective computation that reduces the effective sequence length in later layers while preserving overall behavior.

\subsection{ Overview of DASH}
Motivated by the prefill bottleneck in long-context and multimodal inference, we introduce \textbf{DASH}, a training-free inference-time policy for reducing prefill FLOPs via token-level selective halting in intermediate and deep Transformer blocks, as illustrated in Figure~\ref{fig:overview}.
We consider an input token sequence of length $T$, including both text-only prompts and multimodal prompts formed by concatenating text and visual tokens.
DASH makes no modality-specific assumptions and instead instantiates a unified halting criterion based on a layer-local signal available during the forward pass.
Halting is activated from a start layer $l_s$.
For $l<l_s$, the model processes all $T$ tokens as usual.
At the start layer $l_s$, DASH computes a per-token \(\Delta_{\text{attn}}\) score, ranks tokens by this score, and retains a fixed fraction determined by the pruning ratio $\rho$ to form an active set.
The same active set is then reused for all subsequent layers $l>l_s$, so halted tokens skip both self-attention and feed-forward computation throughout the remaining prefill blocks.

Consequently, the sequence length drops at $l_s$ and remains reduced in deeper layers, reducing the computational cost of both attention and feed-forward sublayers during prefill.

\subsection{\(\Delta_{\text{attn}}\) Signal and Active-Set Update Rule}
Let $\mathbf{H}^{(l)}\in\mathbb{R}^{T\times d}$ denote the token representations entering block $l$ during prefill. We define the attention output \emph{before} the residual addition as
\begin{equation}
\mathbf{U}^{(l)}=\mathrm{Attn}\!\left(\mathrm{LN}(\mathbf{H}^{(l)})\right),
\end{equation}
and compute a layer-local delta-attn score for each token $t$ as
\begin{equation}
\Delta_t^{(l)}=\left\lVert \mathbf{U}^{(l)}_t \right\rVert_2.
\end{equation}
We interpret $\Delta_t^{(l)}$ as a proxy for the marginal benefit of further updating token $t$ at depth $l$.

Given a pruning ratio $\rho\in[0,1)$, DASH performs \emph{single-shot} selection at the start layer $l_s$. Let $S=\{1,\dots,T\}$ be token indices. At layer $l_s$, we select
\begin{equation}
\begin{aligned}
S^\star &= \mathrm{TopK}\!\left(S,\,K,\,\Delta^{(l_s)}\right),\\
K &= \left\lfloor (1-\rho)T \right\rfloor ,
\end{aligned}
\label{eq:active_set}
\end{equation}
reuse $S^\star$ for all subsequent layers, and halt tokens in $S\setminus S^\star$ at $l_s$. Unless stated otherwise, the rule is applied uniformly to all tokens, including multimodal visual tokens when present. Implementation details and pseudo-code are provided in Appendix~\ref{app:ap_algo} (Alg.~\ref{alg:dash}); diagnostic ablations comparing single-shot and multi-shot schedules, and low- versus high-$\Delta_{\mathrm{attn}}$ selection, are reported in Appendix~\ref{sec:appendix_d}.

\definecolor{fullviolet}{RGB}{245,240,250} 
\definecolor{oursviolet}{RGB}{233,221,247} 

\begin{table*}[tb!]
  \centering
  \scriptsize
  \setlength{\tabcolsep}{3.6pt}
  \renewcommand{\arraystretch}{1.10}
    \caption{
    LongBench-E results on \textsc{Qwen2.5-7B-Instruct-1M} under the same prefill-time token reduction setting.
    We compare DASH with representative baselines, including SnapKV adapted to a pruning variant (SnapKV-pr), D$^3$, LLMLingua2, and a long-context adaptation of FastV.
    Avg. (\%) denotes the average score across tasks; higher is better.
    Bold and underlined numbers indicate the best and the second-best results in each column, respectively.
    }
    \vspace{-5pt}
    \label{tab:longbench_acc}
  \begin{tabular}{L A *{13}{Y}}
    \toprule
    \multirow{2}{*}{\textbf{Method}}
    & \multirow{2}{*}{\textbf{Avg. (\%)}}
    & \multicolumn{2}{c}{\textbf{Sing. QA}}
    & \multicolumn{2}{c}{\textbf{Multi. QA}}
    & \multicolumn{2}{c}{\textbf{Summ.}}
    & \multicolumn{3}{c}{\textbf{Few-shot}}
    & \multicolumn{2}{c}{\textbf{Synth. Task}}
    & \multicolumn{2}{c}{\textbf{Code Comp.}} \\
    \cmidrule(lr){3-4}\cmidrule(lr){5-6}\cmidrule(lr){7-8}\cmidrule(lr){9-11}\cmidrule(lr){12-13}\cmidrule(lr){14-15}
    & 
    & \rotatebox{90}{\strut Qasper}
    & \rotatebox{90}{\strut MFQA}
    & \rotatebox{90}{\strut Hotp.QA}
    & \rotatebox{90}{\strut 2Wiki.}
    & \rotatebox{90}{\strut GovRep}
    & \rotatebox{90}{\strut MNews}
    & \rotatebox{90}{\strut TREC}
    & \rotatebox{90}{\strut Triv.QA}
    & \rotatebox{90}{\strut S.Sum}
    & \rotatebox{90}{\strut Count}
    & \rotatebox{90}{\strut Retrieval}
    & \rotatebox{90}{\strut LCC}
    & \rotatebox{90}{\strut Rep.-P} \\
    \midrule

    \textbf{Qwen2.5-7B-Instruct-1M}
    & 48.87
    & 44.19 & 51.13
    & 62.97 & 51.07
    & 6.97  & 6.57
    & 65.00 & 85.75 & 32.95
    & 15.00 & 99.33
    & 59.86 & 54.52 \\
    \midrule
    
    + FastV
    & 43.99
    & \underline{40.44} & 42.63
    & 57.67 & 43.48
    & 6.96  & \underline{6.53}
    & 59.33 & 84.79 & \textbf{33.98}
    & 7.33  & 83.67
    & \underline{52.71} & \underline{52.31} \\
    
    + LLMLingua2
    & 44.16
    & 40.41 & 42.91
    & 51.58 & 43.29
    & 6.44  & 6.41
    & 61.67 & 80.87 & \underline{32.52}
    & 6.33  & \underline{99.00}
    & 49.64 & \textbf{53.05} \\
    
    + D$^3$
    & 45.00
    & 40.18 & \underline{44.49}
    & 60.95 & \textbf{50.16}
    & 6.19  & 6.01
    & \textbf{64.67} & \underline{85.01} & 27.14
    & 15.00 & \textbf{99.33}
    & 45.54 & 40.34 \\

    + SnapKV (pr.)
    & \underline{46.15}
    & 38.14 & 42.98
    & \textbf{61.54} & \underline{48.31}
    & \underline{7.00}  & \textbf{6.70}
    & \underline{63.67} & \textbf{85.25} & 30.87
    & \underline{16.60} & 97.67
    & 51.89 & 49.37 \\

    \rowcolor{gray!10}
    + \textbf{DASH (Ours)}
    & \textbf{46.76}
    & \textbf{40.58} & \textbf{49.38}
    & \underline{61.00} & 48.03
    & \textbf{7.01}  & 6.47
    & 59.00 & 84.21 & 31.82
    & \textbf{16.67} & 98.00
    & \textbf{54.58} & 51.14 \\

    \bottomrule
  \end{tabular}
      \vspace{-10pt}
\end{table*}

\subsection{Execution Semantics, Scheduling, and Complexity}
A token is \emph{halted} once it is excluded from the active set. Halted tokens are removed from the active set, skip both self-attention and feed-forward computation in all subsequent layers, and their hidden states remain fixed at their last updated values. We implement halting by executing each block on the compacted active sequence, so computation scales with the active length rather than the prompt length.

Under the single-shot schedule, layers $0{:}l_s\!-\!1$ process the full prefill sequence of length $T$, while layers $l_s{:}L\!-\!1$ reuse a fixed active set of length $\hat T = \lfloor(1-\rho)T\rfloor$. Using the standard per-layer FLOPs proxy $A(T)$, the baseline and \textsc{DASH} prefill FLOPs are $C_{\text{full}}(T)=L\,A(T)$ and $C_{\text{ours}}(T)=l_sA(T)+(L-l_s)A(\hat T)$, yielding a theoretical prefill FLOPs speedup $s_{\text{FLOPs}}(T)=C_{\text{full}}(T)/C_{\text{ours}}(T)$. For our typical text-only setting on \textsc{Qwen2.5-7B-Instruct-1M} with $l_s=\lfloor 0.4L\rfloor$ and $\rho=0.667$, the theoretical speedup is $1.83\times$ at $T=16{,}384$ tokens. Full derivations and length-dependent results are provided in Appendix~\ref{app:theory_flops_single_shot}.

\section{Experiments}

\label{sec:experiments}

\subsection{Experimental Setup}
\label{sec:exp_setup}

\paragraph{Evaluation focus.}
We evaluate \textsc{DASH} as a \emph{prefill-time} token compression method for long-context inference.
Given an input prompt of length $T$, \textsc{DASH} reduces prefill computation by halting a subset of prompt tokens after a designated start layer $l_s$ and executing subsequent layers only on the compacted active set.
Unless otherwise stated, decoding is kept identical across methods; token compression is applied exclusively to the prompt tokens in prefill.

\paragraph{Backbones.}
For long-context text experiments, we use \textsc{Qwen2.5-7B-Instruct-1M} \citep{qwen251m}.
For vision--language (VL) experiments reported later in \S\ref{sec:vl_results}, we use \textsc{Qwen2-VL-7B} \citep{qwen2vl}.

\paragraph{Benchmarks.}
For long-context text, we consider (i) \textsc{LongBench-E} \citep{longbench}, which covers single-document QA, multi-document QA, summarization, few-shot learning, synthetic tasks, and code completion (Table~\ref{tab:longbench_acc}), and (ii) \textsc{LooGLE} \citep{loogle}, which stresses long-context understanding and includes short/long QA, long summarization, and cloze-style evaluation (Table~\ref{tab:loogle}).
We follow the official evaluation protocols of each benchmark and report their standard task scores; higher is better.

\paragraph{Baselines.}
We compare against representative prefill-time compression baselines:
(i) SnapKV \citep{snapkv} adapted to a token-pruning variant (\textsc{SnapKV(pr.)}),
(ii) \textsc{D$_3$} \citep{d3},
(iii) \textsc{LLMLingua2} \citep{llmlingua2}, and
(iv) a long-context adaptation of \textsc{FastV} \citep{fastv}.
All methods are evaluated under the same backbone and inference configuration.

\paragraph{Operating-point selection (accuracy--speedup trade-off).}
Many baselines expose tunable hyperparameters that induce different accuracy--efficiency trade-offs.
Therefore, in Table~\ref{tab:longbench_acc} and Table~\ref{tab:loogle}, we report each method at a \emph{selected operating point} from its sweep results, chosen to represent its best observed accuracy--speedup trade-off under our evaluation setting.
We report detailed end-to-end quality--efficiency breakdowns separately (Figure~\ref{fig:longbench_acc_e2e}).
For \textsc{DASH}, the sweep is over the start layer $l_s$ and pruning ratio $\rho$ (and any scheduling parameters described in Appendix~\ref{sec:appendix_b}). A lightweight perplexity proxy for screening $l_s$ and the corresponding layer-sweep robustness are detailed in Appendix~\ref{sec:appendix_d}.

\subsection{Long-Context Text Prefill Compression Results}
\label{sec:text_results}

\subsubsection{LongBench-E Main Results}
\label{sec:longbench_results}

Table~\ref{tab:longbench_acc} summarizes \textsc{LongBench-E} results on \textsc{Qwen2.5-7B-Instruct-1M} at the selected operating point for each method.
Among the prefill-time compression baselines, \textsc{DASH} achieves the strongest overall average score (Avg.\ = 46.76), outperforming \textsc{SnapKV(pr.)} (46.15), \textsc{D$^3$} (45.00), \textsc{LLMLingua2} (44.16), and \textsc{FastV} (43.99), while remaining close to the uncompressed backbone (48.87). At this operating point, DASH also achieves a higher  speedup (lower FLOPs) than competing methods at comparable accuracy (Table~\ref{tab:longbench_full}).

Beyond the overall average, \textsc{DASH} remains competitive across diverse task families.
Notably, it is robust on information-intensive settings such as multi-field QA and code completion (e.g., MFQA and LCC in Table~\ref{tab:longbench_acc}), and it preserves strong performance on synthetic retrieval-style tasks where maintaining salient prompt evidence is essential.
These results support that delta-attention-guided halting can reduce prefill computation without disproportionately harming long-context accuracy. These findings generalize to additional text backbones (e.g., Llama-3.1-8B, Qwen-3-8B) and remain consistent under a unified Eager attention kernel.

\paragraph{Kernel compatibility.} Table~\ref{tab:eager_flash} compares DASH under Eager and FlashAttention implementations with identical pruning configurations (pruning ratio = 40\%). Accuracy is essentially unchanged, while FlashAttention provides substantial latency improvement. This confirms that the quality trade-off is governed by the halting rule itself, and that $\Delta_{\mathrm{attn}}$'s kernel compatibility enables additional latency gains without altering selection logic.

\begin{table}[ht]
\small
\centering
\caption{Controlled comparison of Eager and FlashAttention under the same DASH pruning configuration (pruning ratio = 40\%, values in parentheses are speedups).}
\label{tab:eager_flash}
\begin{tabular}{lcc}
\toprule
\textbf{Setting} & \textbf{LongBench-E (Avg)} & \textbf{LooGLE (Avg)} \\
\midrule
Vanilla & 48.87 & 22.69 \\
\midrule
Eager & 46.78 (1.52$\times$) & 19.90 (1.34$\times$) \\
FlashAttn & 46.76 (1.74$\times$) & 19.94 (1.71$\times$) \\
\bottomrule
\end{tabular}
\end{table}

\subsubsection{LooGLE Results}
\label{sec:loogle_results}

LooGLE complements LongBench-E by stressing long-range evidence tracking under QA, long summarization, and cloze-style evaluation. Table~\ref{tab:loogle} reports \textsc{LooGLE} results under the same  setting.

\newcolumntype{L}{>{\raggedright\arraybackslash}p{1.55cm}} 
\newcolumntype{C}{>{\centering\arraybackslash}p{0.73cm}} 
\newcolumntype{N}{>{\centering\arraybackslash}p{0.56cm}}  
\newcolumntype{M}{>{\centering\arraybackslash}p{0.42cm}}
\newcolumntype{T}{>{\raggedright\arraybackslash}m{1.53cm}}

\begin{table}[tb!]
  \centering
  \scriptsize
  \setlength{\tabcolsep}{2.0pt}
  \renewcommand{\arraystretch}{1.08}
  \caption{
    LooGLE results on \textsc{Qwen2.5-7B-Instruct-1M} under the same prefill-time token reduction setting.
    We compare DASH with SnapKV adapted to a token pruning variant (SnapKV pr.), D$^3$, LLMLingua2, and a long-context adaptation of FastV.
    Avg denotes the overall benchmark score; higher is better.
    Bold and underlined numbers indicate the best and the second-best results in each column, respectively.
    }
\vspace{-5pt}
  \label{tab:loogle}
  \begin{tabular}{T C NN NN NN MM}
    \toprule
    \multirow{2}{*}{\textbf{Method}} & \multirow{2}{*}{\textbf{Avg.}}
    & \multicolumn{2}{c}{\textbf{sQA}}
    & \multicolumn{2}{c}{\textbf{lQA}}
    & \multicolumn{2}{c}{\textbf{lSum}}
    & \multicolumn{2}{c}{\textbf{sCloze}} \\
    \cmidrule(lr){3-4}\cmidrule(lr){5-6}\cmidrule(lr){7-8}\cmidrule(lr){9-10}
     &  & \textbf{BF1} & \textbf{RL}
        & \textbf{BF1} & \textbf{RL}
        & \textbf{BF1} & \textbf{RL}
        & \textbf{Ex.} & \textbf{Part.} \\
    \midrule

    \textbf{Qwen2.5-7B-Instruct-1M}
    & 22.69
    & 0.842 & 0.234
    & 0.830 & 0.123
    & 0.842 & 0.043
    & 73.5  & 83.3 \\
    \midrule

    + D$^3$
    & 19.49
    & 0.830 & 0.124
    & 0.822 & 0.078
    & 0.831 & 0.038
    & 72.1  & 81.1 \\

    + LLMLingua2
    & 19.56
    & \textbf{0.841} & \underline{0.180}
    & \underline{0.828} & \underline{0.113}
    & \textbf{0.842} & \textbf{0.049}
    & \underline{73.1} & 80.5 \\

    + FastV
    & 19.78
    & \underline{0.840} & \textbf{0.202}
    & \textbf{0.829} & 0.106
    & 0.840 & \underline{0.040}
    & 73.0  & \underline{82.4} \\

    + SnapKV (pr.)
    & \underline{19.87}
    & 0.836 & 0.171
    & 0.827 & \textbf{0.116}
    & \underline{0.841} & \underline{0.043}
    & \underline{73.2} & \underline{82.9} \\

    \rowcolor{gray!10}
    + \textbf{DASH (Ours)}
    & \textbf{19.94}
    & 0.839 & 0.172
    & \textbf{0.829} & 0.112
    & \underline{0.841} & \underline{0.040}
    & \textbf{73.6} & \textbf{83.1} \\

    \bottomrule
  \end{tabular}

\end{table}

\textsc{DASH} delivers the best overall score among the compared compression methods (Avg.\ = 19.94), slightly exceeding \textsc{SnapKV(pr.)} (19.87) and improving over \textsc{FastV}, \textsc{LLMLingua2}, and \textsc{D3}.
On cloze-style evaluation, \textsc{DASH} attains the highest Exact and Partial scores (73.6 and 83.1, respectively), indicating that it better preserves the long-context evidence required for precise span-level inference.
Overall, the \textsc{LooGLE} results further corroborate that \textsc{DASH} maintains long-context understanding under prefill-time token reduction.

\subsection{Vision--Language Prefill Compression Results}
\label{sec:vl_results}

\begin{figure}[tb!]
    \centering
    \includegraphics[width=\columnwidth]{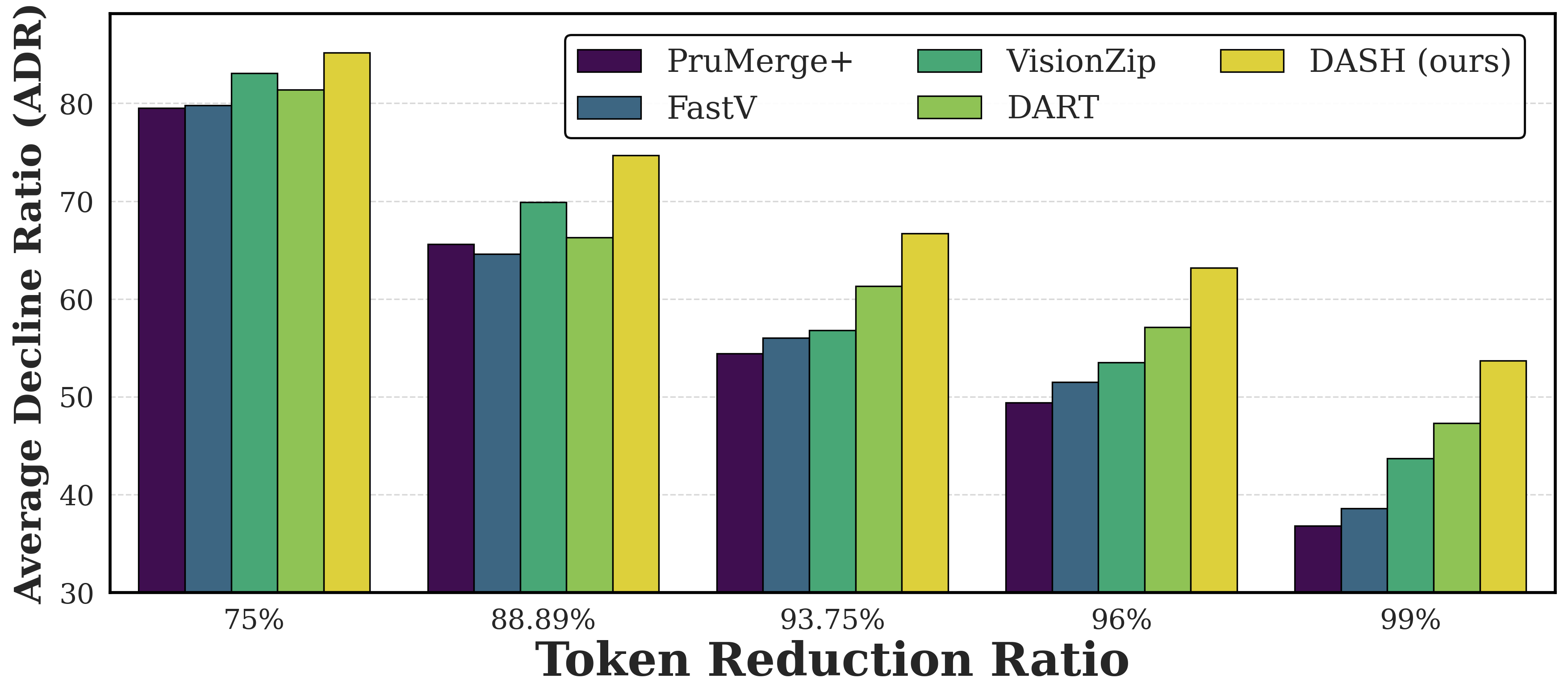}
    \vspace{-15pt}
    \caption{Vision--language token compression on Qwen2-VL-7B across six vision--language benchmarks. We compare PruMerge+, FastV, VisionZip, DART, and DASH (Ours) under different token reduction ratios. Performance is reported as the average decline ratio (ADR), averaged over benchmarks, with the uncompressed model normalized to 100.0. Full per-benchmark results are provided in Table~\ref{tab:vl_full} in the appendix.}
    \label{fig:vl_acc}
\end{figure}

Figure~\ref{fig:vl_acc} summarizes vision--language token compression results on \textsc{Qwen2-VL-7B} across six VL benchmarks under varying token reduction ratios.
We report performance using the average decline ratio (ADR), where the uncompressed model is normalized to $100.0$ and higher values indicate better retention of task accuracy.

Across all reduction ratios, \textsc{DASH} consistently achieves the highest ADR among compared methods. Notably, its advantage becomes pronounced under aggressive compression.
At moderate reduction (e.g., $75\%$--$88.9\%$), \textsc{DASH} already matches or exceeds prior methods, while at extreme reduction levels ($96\%$ and $99\%$), it exhibits substantially slower degradation than \textsc{FastV}, \textsc{VisionZip}, and \textsc{DART}.
This trend indicates delta-attention-guided halting is particularly effective in identifying and preserving the small subset of visual--textual tokens that remain influential in deeper layers.

This robustness is consistent with two observations.
First, VL prompts typically contain a large number of visually redundant tokens whose contributions saturate early, making them suitable candidates for halting. Second, by relying on attention-branch deltas rather than heuristic scores, \textsc{DASH} may better capture cross-modal interactions that remain active across layers.
Full per-benchmark results are provided in Table~\ref{tab:vl_full} in the appendix. Extended evaluations on stronger VL backbones (Qwen2.5-VL-7B and InternVL-2.5-8B) and additional baselines are also reported in Appendix~\ref{sec:appendix_d}.

\begin{figure}[tb!]
    \centering
    \includegraphics[width=0.85\columnwidth]{figs/exp_e2e4.png}
    \caption{Overall LongBench-E performance trade-off between end-to-end (E2E) time and score. DASH achieves higher accuracy at lower latency, outperforming FastV by 8.5\% in score and running 1.74× faster than the baseline under comparable accuracy.}
    \vspace{-10pt}
    \label{fig:longbench_acc_e2e}
\end{figure}

\begin{figure*}[tb!]
    \centering
    \includegraphics[width=2\columnwidth]{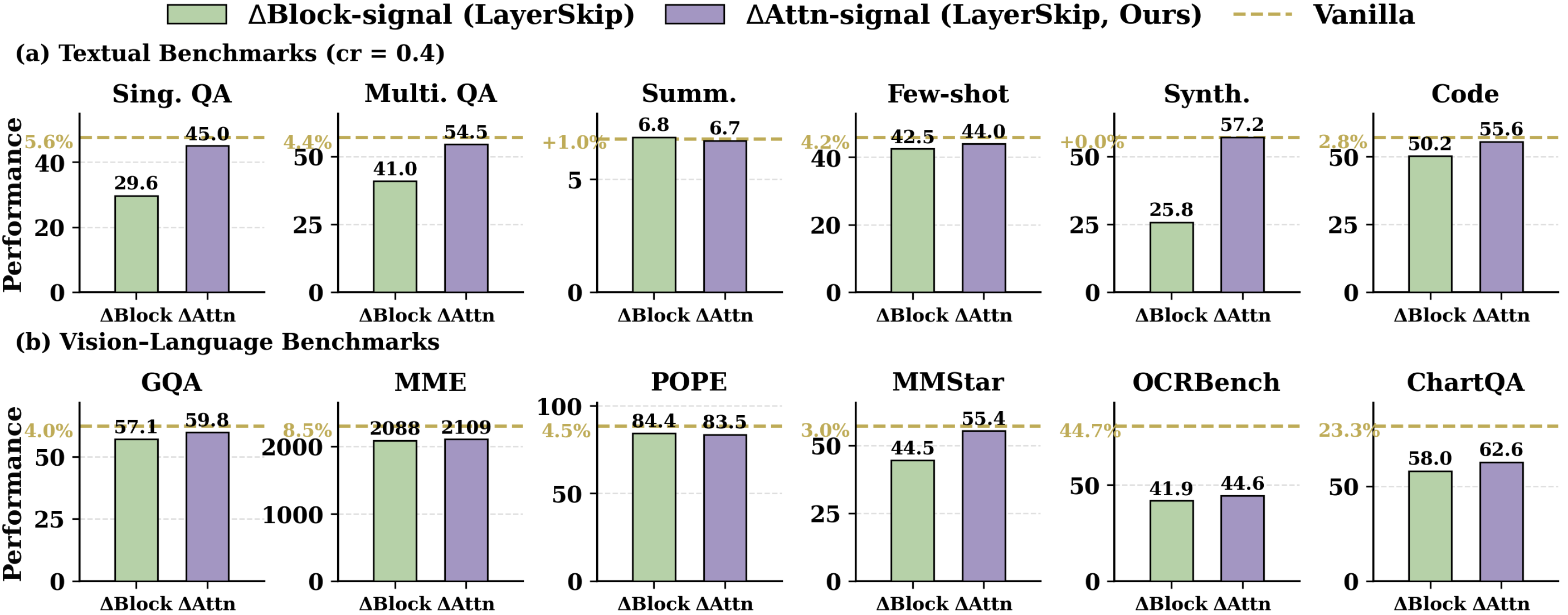}
    \caption{
    Ablation of delta-signal choices for layer-level token halting.
    \textbf{(a)} Textual benchmarks from \textsc{LongBench-E} at compression ratio $\mathrm{cr}=0.4$, evaluated on \textsc{Qwen2.5-7B-Instruct-1M}.
    \textbf{(b)} Vision--language benchmarks evaluated on \textsc{Qwen2-VL-7B} with a fixed token removal ratio of $75\%$.
    Bars compare \textit{$\Delta$Block-signal} (layer-level delta from the whole Transformer block) and
    \textit{$\Delta$Attn-signal} (delta from the attention branch only), where both variants apply the same token halting mechanism and differ only in the delta signal used for decision making.
    The dashed horizontal line denotes the \textit{Vanilla} (no layer skipping) baseline performance for each benchmark,
    and the percentage annotated above it indicates the relative performance change of the best layer-skipping variant
    with respect to the Vanilla baseline.
    }
    \label{fig:ablation}
\end{figure*}

\subsection{End-to-End Efficiency and Accuracy Trade-off}
\label{sec:e2e_tradeoff}

While the above results focus on task accuracy at selected operating points, prefill-time compression ultimately aims to improve end-to-end (E2E) inference efficiency.
Figure~\ref{fig:longbench_acc_e2e} plots the trade-off between overall \textsc{LongBench-E} score and E2E inference time. Because prefill dominates time-to-first-token in long-context settings, reductions in prefill FLOPs translate directly into end-to-end latency improvements.

\textsc{DASH} occupies a favorable region of the trade-off curve, achieving higher accuracy at lower latency than all baselines.
Under comparable E2E time, it outperforms \textsc{FastV} by $8.5\%$.
Conversely, at comparable accuracy to the uncompressed baseline, \textsc{DASH} runs $1.74\times$ faster with only a marginal performance drop.
These gains stem primarily from reduced prefill computation, while decoding remains unchanged across methods.

Importantly, these improvements are not achieved by trading accuracy for speed in a narrow regime.
Instead, \textsc{DASH} offers a smooth accuracy--efficiency frontier, allowing practitioners to select operating points that balance latency and quality according to deployment constraints.
A detailed breakdown of task-wise E2E quality and efficiency metrics is reported in Appendix Figure~\ref{fig:longbench_acc_e2e}.

\begin{figure*}[tb!]
  \centering
  \begin{minipage}[t]{0.48\textwidth}
    \centering
    \includegraphics[width=\linewidth]{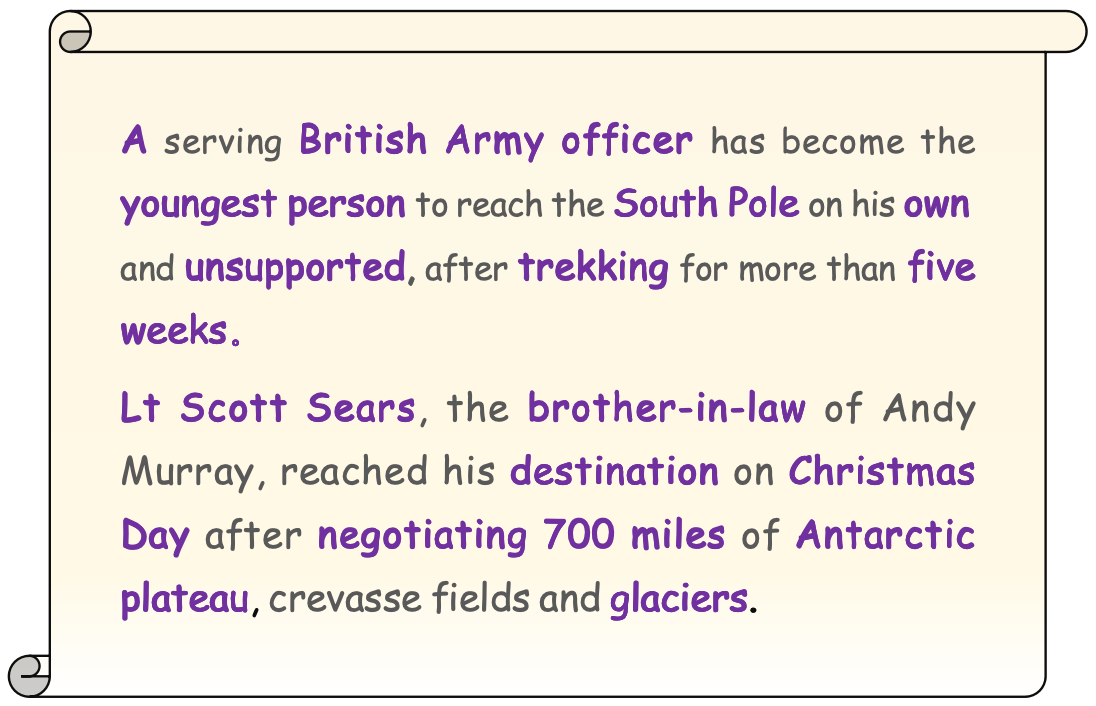}
  \end{minipage}\hfill
  \begin{minipage}[t]{0.48\textwidth}
    \centering
    \includegraphics[width=\linewidth]{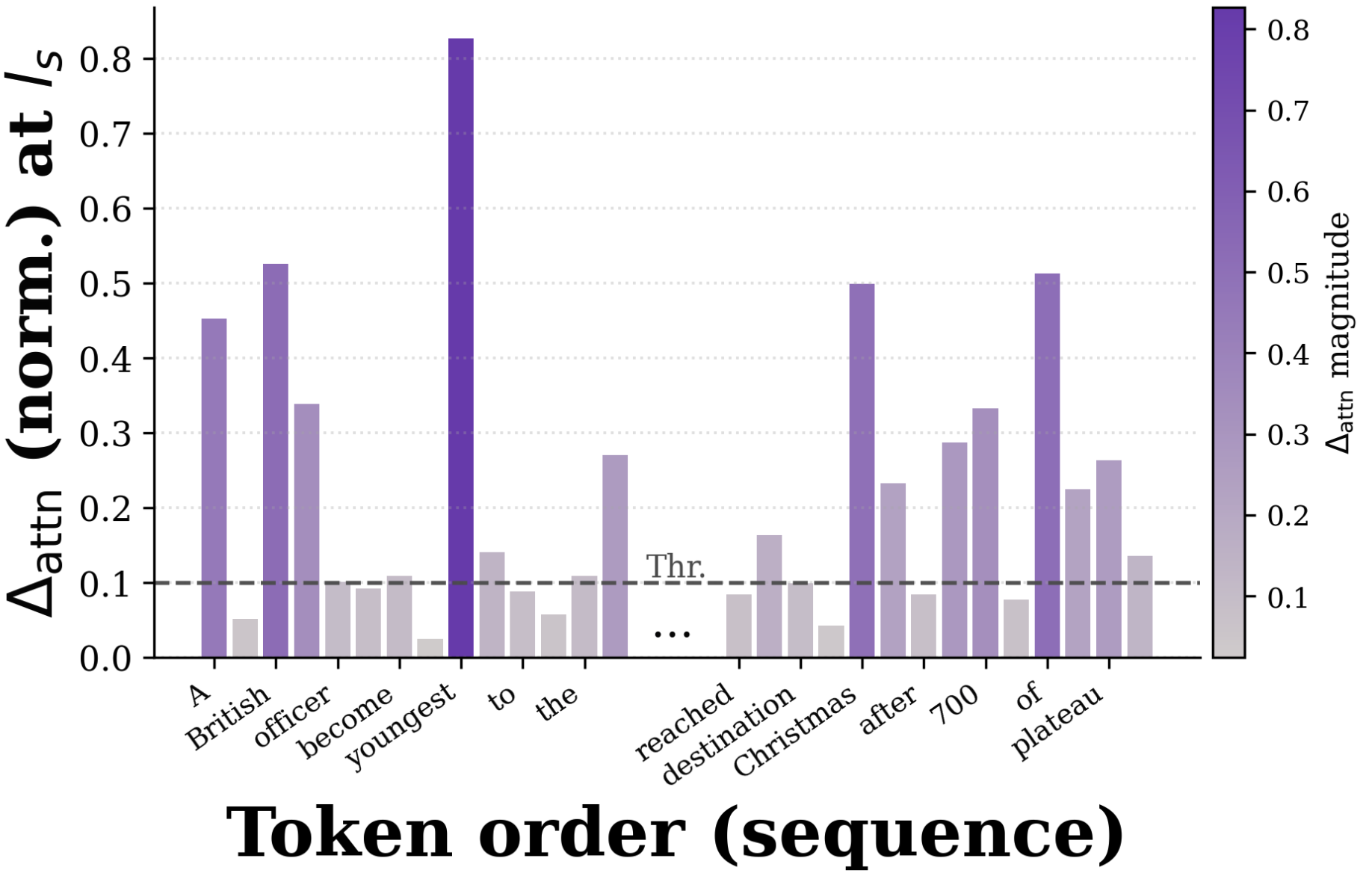}
  \end{minipage}
  \caption{Qualitative example from LongBench-E (MultiNews) showing that token-wise $\Delta_{\mathrm{attn}}$ at the decision layer $l_s$ tends to assign higher scores to semantically informative tokens. The dashed line indicates the pruning threshold and ellipses denote omitted tokens.}
  \label{fig:disc_tokens}
  \vspace{-5pt}
\end{figure*}

\subsection{Ablation: Choice of Delta Signal for Token Halting}
\label{sec:ablation}

We ablate the choice of delta signal used for token-level halting applied from layer $l_s$.
Specifically, we compare \textit{$\Delta_{\mathrm{block}}$-signal}, which measures representation change across the entire Transformer block, against \textit{$\Delta_{\mathrm{attn}}$-signal}, which isolates the contribution from the attention branch.
Both variants employ identical layer-skipping mechanisms and differ only in the signal used for halting decisions.

As shown in Figure~\ref{fig:ablation}(a), on textual benchmarks from \textsc{LongBench-E} at compression ratio $\mathrm{cr}=0.4$, $\Delta_{\mathrm{attn}}$-signal consistently outperforms $\Delta_{\mathrm{block}}$-signal across most task categories, including single- and multi-document QA, few-shot learning, synthetic tasks, and code completion, while maintaining comparable summarization performance.
A similar pattern is observed on vision--language benchmarks (Figure~\ref{fig:ablation}(b)), where $\Delta_{\mathrm{attn}}$-signal yields higher accuracy on tasks such as GQA, MMStar, OCRBench, and ChartQA.

These results suggest that attention-branch deltas provide a more faithful indicator of token relevance in deeper layers.
While block-level deltas conflate attention updates with feed-forward transformations that may reflect local reparameterization rather than contextual influence, attention deltas more directly capture whether a token continues to participate in global information aggregation.
This property makes $\Delta_{\mathrm{attn}}$-signal better suited for guiding token halting under aggressive  compression.

\paragraph{Directionality of $\Delta_{\mathrm{attn}}$.} Table~\ref{tab:direction_ablation} ablates the directionality of the halting criterion under the same pruning budget (pruning ratio = 40\%). The large gap between low-$\Delta_{\mathrm{attn}}$ (our default) and both high-$\Delta_{\mathrm{attn}}$ and random selection indicates that the directionality of $\Delta_{\mathrm{attn}}$ is crucial, supporting the interpretation that stabilized tokens are indeed redundant.

\begin{table}[ht]
\small
\centering
\caption{Directional ablation: low-$\Delta_{\mathrm{attn}}$ vs. high-$\Delta_{\mathrm{attn}}$ vs. random halting (pruning ratio = 40\%).}
\label{tab:direction_ablation}
\resizebox{0.48\textwidth}{!}{
\begin{tabular}{lcc}
\toprule
\textbf{Method} & \textbf{LongBench-E (Avg.)} & \textbf{LooGLE (Avg.)} \\
\midrule
Vanilla & 48.87 & 22.69 \\
\midrule
+ Random & 33.65 & 10.16 \\
+ High $\Delta_{\mathrm{attn}}$ & 25.45 & 10.22 \\
\rowcolor{gray!10}
+ \textbf{Low $\Delta_{\mathrm{attn}}$ (Ours)} & \textbf{46.76} & \textbf{19.94} \\
\bottomrule
\end{tabular}}
\end{table}

\section{Discussion}
\label{sec:discussion}

\paragraph{Token dynamics: a heavy-tailed view from $\Delta_{\mathrm{attn}}$.}
Figure~\ref{fig:disc_tokens} provides an intuitive explanation of why \textsc{DASH} can aggressively reduce prefill computation while retaining accuracy.
At the decision layer $l_s$, token-wise $\Delta_{\mathrm{attn}}$ exhibits a pronounced heavy-tailed pattern: most tokens lie in a narrow low-$\Delta_{\mathrm{attn}}$ band, while only a small fraction forms a long tail with substantially larger updates.
This structure naturally supports a natural separation for single-shot TopK selection at $l_s$.
Tokens with small $\Delta_{\text{attn}}$ at $l_s$ are unlikely to benefit from further updates, so they can be halted at $l_s$ with minor impact, while tail tokens are retained in the active set.
In this sense, $\Delta_{\mathrm{attn}}$ acts as a layer-local indicator of whether a token remains \emph{active} in shaping attention updates beyond $l_s$.

\paragraph{Why the attention-branch delta matters.}
The ablation in Figure~\ref{fig:ablation} further clarifies that not all delta signals are equally suitable for halting decisions.
Using $\Delta_{\mathrm{attn}}$-signal consistently outperforms the block-level alternative across both long-context text and vision--language benchmarks under the same layer-skipping mechanism.
A plausible interpretation is that attention-branch deltas directly reflect a token's marginal participation in cross-token (and cross-modal) aggregation, whereas block-level deltas conflate attention updates with feed-forward transformations that may alter representations without indicating continued global influence.
Combined with the heavy-tailed dynamics observed in Figure~\ref{fig:disc_tokens}, this explains why $\Delta_{\mathrm{attn}}$ yields a more reliable separation between tokens that can be halted and tokens that should remain in the active set. Appendix~\ref{sec:appendix_d} further verifies that $\Delta_{\mathrm{attn}}$ is a faithful proxy for full-attention importance ranking. The directional ablation in Table~\ref{tab:direction_ablation} corroborates this interpretation: halting tokens with low $\Delta_{\mathrm{attn}}$ preserves accuracy, whereas reversing the criterion or selecting at random causes severe degradation.

\paragraph{Implications for robustness under aggressive compression.}
The heavy-tailed separation in Figure~\ref{fig:disc_tokens} also sheds light on the empirical trend that \textsc{DASH} degrades more gracefully at higher reduction ratios.
When compression becomes aggressive, the primary failure mode is to mistakenly discard the small set of tokens that carry persistent global influence.
A signal with clearer tail separation is therefore crucial in the high-compression regime.
This aligns with the vision--language results in Figure~\ref{fig:vl_acc}, where \textsc{DASH} maintains a larger margin over prior methods as token reduction increases.

\paragraph{From prefill savings to end-to-end gains.}
Finally, Figure~\ref{fig:longbench_acc_e2e} shows that the token-level decisions guided by $\Delta_{\mathrm{attn}}$ translate into a favorable end-to-end accuracy--latency frontier.
Because \textsc{DASH} targets prefill-time computation while keeping decoding identical, improvements in prefill efficiency can be directly reflected in reduced E2E time at comparable quality.
Together, the qualitative evidence in Figure~\ref{fig:disc_tokens} and the ablation in Figure~\ref{fig:ablation} provide a coherent explanation for why \textsc{DASH} achieves strong accuracy retention (Tables~\ref{tab:longbench_acc}, \ref{tab:loogle}) while improving practical serving efficiency.

\section{Conclusion}
\label{sec:conclusion}

We introduced \textsc{DASH}, a training-free inference-time method that reduces long-context prefill computation via delta-attention-guided selective token halting: \textsc{DASH} selects a compact active set at a start layer $l_s$ using token-wise $\Delta_{\mathrm{attn}}$ and reuses it in subsequent layers, while keeping decoding unchanged. Our analyses further show that attention-branch deltas are more reliable than block-level deltas for halting decisions, and token-level visualizations reveal a heavy-tailed $\Delta_{\mathrm{attn}}$ pattern that separates a small set of persistently influential tokens from largely redundant ones. Overall, \textsc{DASH} offers a practical and effective approach to accelerating prefill for long-context and multimodal inference, making AI applications affordable.

\clearpage
\section*{Limitations}
This paper focuses on the core design and evaluation of \(\Delta_{\text{attn}}\)–guided selective token halting for prefill acceleration. While we benchmark DASH on a broad set of long-context and vision–language tasks, some additional ablations and per-setting results are omitted from the main text due to space constraints and can be included in supplementary material. Moreover, we do not exhaustively explore all model scales, architectures, or deployment configurations; studying these extensions and their interactions with system-level optimizations is left for future work.

\section*{Acknowledgments}
This work was supported by New Generation Artificial Intelligence-National Science and Technology Major Project.

\bibliography{main}
\newpage

\appendix

\section{Theoretical FLOPs}
\label{sec:appendix_a}
\subsection{Theoretical FLOPs of Single-shot Halting in DASH}
\label{app:theory_flops_single_shot}

\paragraph{FLOPs proxy.}
Following FastV~\citep{fastv}, we approximate the per-layer FLOPs of a Transformer block (MHA + FFN) for sequence length $n$ as
\begin{equation}
A(n) \;=\; 4nd^2 + 2n^2d + 2ndm,
\label{eq:perlayer_flops_fastv}
\end{equation}
where $d$ is the hidden size and $m$ is the FFN intermediate size.

\paragraph{Single-shot halting.}
Let $L$ be the number of layers and $l_s$ the halting start layer. Under single-shot halting, layers $0\!:\!l_s\!-\!1$ process the full sequence of length $n$, while layers $l_s\!:\!L\!-\!1$ reuse a fixed active set with length $\hat n$.

\paragraph{Effective kept length.}
All FLOPs analysis in this section focuses on the prefill stage; decoding-stage costs are unaffected by single-shot halting and thus excluded. Our implementation always keeps a prefix of $n_{\text{first}}$ tokens and a suffix of $n_{\text{last}}$ tokens, i.e., $n_{\text{fix}}=n_{\text{first}}+n_{\text{last}}$, and drops a fraction $c$ of the remaining eligible tokens. Thus,
\begin{equation}
\hat n \;=\; n_{\text{fix}} + (1-c)(n-n_{\text{fix}})
\;=\; (1-c)n + c\,n_{\text{fix}}.
\label{eq:nkeep}
\end{equation}

\paragraph{Total FLOPs, reduction, and speedup.}
The baseline and single-shot FLOPs are
\begin{align}
C_{\text{full}}(n) &= L\,A(n), \label{eq:full_flops}\\
C_{\text{ours}}(n) &= l_sA(n) + (L-l_s)A(\hat n). \label{eq:ours_flops}
\end{align}
Therefore,
\begin{align}
r_{\text{FLOPs}}(n)
&= 1 - \frac{C_{\text{ours}}(n)}{C_{\text{full}}(n)} \nonumber \\
&= 1 - \frac{l_sA(n) + (L-l_s)A(\hat n)}{L\,A(n)},
\label{eq:flops_reduction}
\\
s_{\text{FLOPs}}(n)
&= \frac{C_{\text{full}}(n)}{C_{\text{ours}}(n)}.
\label{eq:flops_speedup}
\end{align}
Since $A(n)$ contains both $O(n)$ and $O(n^2)$ terms, $s_{\text{FLOPs}}(n)$ depends on the input length $n$.

\paragraph{Instantiation (Qwen2.5-7B).}
We use $L{=}28$, $d{=}3584$, $m{=}18944$, and $l_s{=}\lfloor 0.4L\rfloor{=}11$.
With $n_{\text{first}}{=}64$, $n_{\text{last}}{=}32$ ($n_{\text{fix}}{=}96$) and $c{=}0.667$, Eq.~\eqref{eq:nkeep} gives
\begin{equation}
\hat n \;=\; (1-0.667)n + 0.667 \cdot 96.
\end{equation}

\paragraph{Length dependence.}
Because the attention term in $A(n)$ scales as $O(n^2)$, the FLOPs reduction and speedup are length-dependent.
Table~\ref{tab:theory_speedup_vs_len} reports $r_{\text{FLOPs}}(n)$ and $s_{\text{FLOPs}}(n)$ for representative text-only prefill lengths; notably, at $n{=}16{,}384$ the theoretical speedup is $1.83\times$, matching our typical measured setting.

\begin{table}[!t]
\centering
\small
\setlength{\tabcolsep}{6pt}
\caption{Theoretical FLOPs reduction and speedup under single-shot halting for different text-only prefill lengths $n$.
We follow the per-layer FLOPs proxy $A(n)$ in Eq.~\eqref{eq:perlayer_flops_fastv} and compute $r_{\text{FLOPs}}(n)$ and $s_{\text{FLOPs}}(n)$ using Eq.~\eqref{eq:flops_reduction}--\eqref{eq:flops_speedup}.
We instantiate Qwen2.5-7B-Instruct-1M with $L{=}28$, $l_s{=}\lfloor 0.4L\rfloor{=}11$, and our implementation parameters $\texttt{compression\_ratio}=0.667$, $\texttt{keep\_first\_n}=64$, $\texttt{keep\_last\_n}=32$. The kept length $\hat n$ accounts for the fixed prefix/suffix and the rounding in $\texttt{drop\_target}=\lfloor \mathrm{round}(c\cdot (n-n_{\mathrm{fix}}))\rfloor$, where $n_{\mathrm{fix}}=96$. All FLOPs and speedup numbers correspond to the prefill stage.}
\label{tab:theory_speedup_vs_len}
\begin{tabular}{l c c c}
\toprule
$n$ & $\hat n$ (kept) & $r_{\text{FLOPs}}(n)$ & $s_{\text{FLOPs}}(n)$ \\
\midrule
8{,}192   & 2{,}792  & 43.28\% & 1.76$\times$ \\
16{,}384  & 5{,}520  & 45.49\% & 1.83$\times$ \\
32{,}768  & 10{,}976 & 47.90\% & 1.92$\times$ \\
65{,}536  & 21{,}888 & 50.09\% & 2.00$\times$ \\
131{,}072 & 43{,}711 & 51.72\% & 2.07$\times$ \\
\bottomrule
\end{tabular}
\end{table}

\section{Algorithm Pseudocode}
\label{sec:appendix_b}

\label{app:ap_algo}
For completeness, we provide the pseudocode of \textbf{DASH} used in the prefill stage. The algorithm performs a single-shot token selection at the start layer $l_s$ based on the per-token delta-attn score (the $\ell_2$ norm of the pre-residual attention output), and reuses the resulting active set for all subsequent layers under a fixed pruning ratio $\rho$.
\begin{algorithm}[t]
\small
\caption{DASH for prefill token halting with a fixed pruning ratio}
\label{alg:dash}
\begin{algorithmic}[1]
\STATE \textbf{Input:} token sequence of length $T$ (text-only or multimodal), Transformer with $L$ blocks, start layer $l_s$, pruning ratio $\rho$
\STATE \textbf{Initialize:} $S_1 \leftarrow \{1,\dots,T\}$ and $T_1 \leftarrow T$
\FOR{$l=1$ \TO $L$}
    \STATE \textbf{(Active tokens)} form $\mathbf{H}^{(l)} \in \mathbb{R}^{T_l \times d}$ by gathering tokens indexed by $S_l$
    \STATE \textbf{(Pre-residual attention output)} $\mathbf{U}^{(l)} \leftarrow \mathrm{Attn}\!\left(\mathrm{LN}(\mathbf{H}^{(l)})\right)$
    \STATE \textbf{(Block forward)} $\widetilde{\mathbf{H}}^{(l)} \leftarrow \mathbf{H}^{(l)} + \mathbf{U}^{(l)}$
    \STATE $\mathbf{H}^{(l+1)} \leftarrow \widetilde{\mathbf{H}}^{(l)} + \mathrm{FFN}\!\left(\mathrm{LN}(\widetilde{\mathbf{H}}^{(l)})\right)$
    \IF{$l = l_s$}
    \STATE $\Delta_t^{(l_s)} \leftarrow \lVert \mathbf{U}^{(l_s)}_t \rVert_2$ for $t \in S$
    \STATE $K \leftarrow \lfloor (1-\rho)T \rfloor$
    \STATE $S^\star \leftarrow \mathrm{TopK}(S, K, \Delta^{(l_s)})$
    \ENDIF
\STATE \textbf{if} $l > l_s$ \textbf{then} use $S^\star$ as the active index set
\ENDFOR
\end{algorithmic}
\end{algorithm}

\section{More Experimental Results}
\label{sec:appendix_c}
\subsection{Long-context Language Benchmarks: Detailed Settings and Results}
\label{app:text_details}

\paragraph{Backbone and benchmarks.}
We evaluate long-context text inference on Qwen2.5-7B-Instruct-1M across LongBench-E and LooGLE, following the standard evaluation protocol of each benchmark. We focus on \emph{prefill-time} token halting and report both task performance and efficiency metrics.

\paragraph{DASH settings.}
All texual experiments are conducted on Nvidia GPUs. Qwen2.5-7B-Instruct-1M is evaluated on long-text benchmarks using A100
(40GB). DASH is activated from a fixed start layer $l_s=11$ for all text experiments. At layer $l_s$, we compute token-wise $\Delta_{\mathrm{attn}}$ and select a compact active set by retaining a fraction $(1-\rho)$ of tokens (with the same token reduction ratio definition used throughout the paper). The active set is reused for all subsequent layers $l>l_s$ during prefill. We do not prune generated tokens; however, halting prompt tokens shortens the KV cache and can also reduce the per-token decoding cost.

\paragraph{Baselines and text adaptations.}
We compare DASH with representative token-compression baselines. For \textsc{FastV}, we use a text-adapted variant: since vanilla FastV is designed for vision--language inputs and typically selects tokens based on vision-heavy attention patterns, we instead derive the retention mask from a text-only attention proxy (computed at the designated prune layer) and apply the same single-shot keep mask to all deeper prefill layers. This yields a comparable \emph{prefill-time} token pruning policy for text without requiring access to full attention matrices under FlashAttention/SDPA.

For \textsc{SnapKV}, we adopt a token-pruning adaptation to ensure a fair comparison in the \emph{prefill-compute} regime: rather than post-hoc KV compression after computing all tokens, we convert SnapKV's selection signal into a single-shot keep mask at the chosen layer and physically reduce the sequence for subsequent layers (i.e., dropped tokens skip both attention and FFN computation in deeper prefill blocks). This adaptation isolates the effect of token dropping on prefill FLOPs and latency under the same token reduction ratio.

\paragraph{Vision experiments.}
All experiments are conducted on Nvidia GPUs. 
Qwen2-VL-7B-Instruct is evaluated on image benchmarks using RTX A6000 (48GB). For ToMe and ToFu, we follow the original implementations with a minor modification: tokens are reduced proportionally across layers until the target sparsity is reached. Other baseline settings follow the original papers.

\begin{table*}[!t]
  \small
  \centering
  \scriptsize
  \setlength{\tabcolsep}{3.5pt}
  \renewcommand{\arraystretch}{1.15}
  \caption{LongBench-E quality (Score) and efficiency relative to Qwen2.5-7B-Instruct-1M, including prefill, decoding, end-to-end latency, throughput, and peak GPU memory. \textbf{Bold} denotes the best compressed-model result, and \underline{underline} denotes the second best (excluding the original model).}
  \label{tab:longbench_full}
  \resizebox{\textwidth}{!}{%
  \begin{tabular}{l|rrr|r|rrrrr}
    \toprule
    & \multicolumn{4}{c|}{\textbf{Score}} & \multicolumn{5}{c}{\textbf{Efficiency vs. Full} ($\uparrow$)} \\
    \cmidrule(lr){2-5}\cmidrule(lr){6-10}
    \textbf{Method} & \textbf{D1} & \textbf{D2} & \textbf{D3} & \textbf{Avg} & \textbf{Prefill} & \textbf{Gen.} & \textbf{Total} & \textbf{Thrpt.} & \textbf{Mem.} \\
    \midrule

    \multicolumn{10}{l}{\textbf{Single-Document QA} \textit{(Datasets: qasper\_e, multifieldqa\_en\_e)}} \\
    \midrule
    Qwen2.5-7B-Instruct-1M & 44.19 & 51.13 & -- & 47.66 & 1.00$\times$ & 1.00$\times$ & 1.00$\times$ & 1.00$\times$ & 1.00$\times$ \\
    + LLMLingua2 & 40.41 & 42.91 & -- & 41.66 & \underline{1.24$\times$} & 0.74$\times$ & 0.77$\times$ & 0.77$\times$ & 0.92$\times$ \\
    + SnapKV (pr.) & 38.14 & 42.98 & -- & 40.56 & 1.23$\times$ & \underline{1.37$\times$} & \underline{1.26$\times$} & \underline{1.26$\times$} & \underline{1.01$\times$} \\
    + D\textsuperscript{3} & 40.18 & 44.49 & -- & 42.33 & 1.14$\times$ & 1.24$\times$ & 1.12$\times$ & 1.12$\times$ & 1.00$\times$ \\
    \rowcolor{gray!10}
    + \textbf{DASH (Ours)} & 40.58 & 49.38 & -- & \textbf{44.98} & \textbf{1.72$\times$} & \textbf{1.56$\times$} & \textbf{1.56$\times$} & \textbf{1.56$\times$} & \textbf{1.02$\times$} \\
    \midrule 

    \multicolumn{10}{l}{\textbf{Multi-Document QA} \textit{(Datasets: hotpotqa\_e, 2wikimqa\_e)}} \\
    \midrule
    Qwen2.5-7B-Instruct-1M & 62.97 & 51.07 & -- & 57.02 & 1.00$\times$ & 1.00$\times$ & 1.00$\times$ & 1.00$\times$ & 1.00$\times$ \\
    + LLMLingua2 & 51.58 & 43.29 & -- & 47.44 & \underline{1.32$\times$} & 0.80$\times$ & 0.73$\times$ & 0.73$\times$ & 0.92$\times$ \\
    + SnapKV (pr.) & 61.54 & 48.31 & -- & \underline{54.93} & 1.30$\times$ & \underline{1.51$\times$} & \underline{1.18$\times$} & \underline{1.18$\times$} & \underline{1.01$\times$} \\
    + D\textsuperscript{3} & 60.95 & 50.16 & -- & \textbf{55.56} & 1.17$\times$ & 1.34$\times$ & 1.00$\times$ & 1.00$\times$ & 1.00$\times$ \\
    \rowcolor{gray!10}
    + \textbf{DASH (Ours)} & 61.00 & 48.03 & -- & 54.52 & \textbf{1.74$\times$} & \textbf{1.74$\times$} & \textbf{1.45$\times$} & \textbf{1.45$\times$} & \textbf{1.02$\times$} \\
    \midrule

    \multicolumn{10}{l}{\textbf{Summarization} \textit{(Datasets: gov\_report\_e, multi\_news\_e)}} \\
    \midrule
    Qwen2.5-7B-Instruct-1M & 6.97 & 6.57 & -- & 6.77 & 1.00$\times$ & 1.00$\times$ & 1.00$\times$ & 1.00$\times$ & 1.00$\times$ \\
    + LLMLingua2 & 6.44 & 6.41 & -- & 6.42 & \underline{1.47$\times$} & 0.74$\times$ & 0.80$\times$ & 0.80$\times$ & 0.96$\times$ \\
    + SnapKV (pr.) & 7.00 & 6.70 & -- & \textbf{6.85} & 1.27$\times$ & \underline{1.18$\times$} & \underline{1.11$\times$} & \underline{1.11$\times$} & \underline{1.03$\times$} \\
    + D\textsuperscript{3} & 6.19 & 6.01 & -- & 6.10 & 1.26$\times$ & 1.10$\times$ & 1.06$\times$ & 1.06$\times$ & 1.00$\times$ \\
    \rowcolor{gray!10}
    + \textbf{DASH (Ours)} & 7.01 & 6.47 & -- & \underline{6.74} & \textbf{1.89$\times$} & \textbf{1.19$\times$} & \textbf{1.28$\times$} & \textbf{1.28$\times$} & \textbf{1.03$\times$} \\
    \midrule

    \multicolumn{10}{l}{\textbf{Few-shot Learning} \textit{(Datasets: trec\_e, triviaqa\_e, samsum\_e)}} \\
    \midrule
    Qwen2.5-7B-Instruct-1M & 65.00 & 85.75 & 32.95 & 45.93 & 1.00$\times$ & 1.00$\times$ & 1.00$\times$ & 1.00$\times$ & 1.00$\times$ \\
    + LLMLingua2 & 61.67 & 80.87 & 32.53 & 43.77 & \underline{1.47$\times$} & 0.80$\times$ & 0.79$\times$ & 0.79$\times$ & 0.97$\times$ \\
    + SnapKV (pr.) & 63.67 & 85.25 & 30.87 & \textbf{44.95} & 1.21$\times$ & 1.21$\times$ & 1.03$\times$ & 1.03$\times$ & \textbf{1.03$\times$} \\
    + D\textsuperscript{3} & 64.67 & 85.01 & 27.14 & \underline{44.21} & 1.21$\times$ & 1.19$\times$ & 1.01$\times$ & 1.01$\times$ & 1.00$\times$ \\
    \rowcolor{gray!10}
    + \textbf{DASH (Ours)} & 59.00 & 84.21 & 31.82 & 43.76 & \textbf{1.82$\times$} & \textbf{1.53$\times$} & \textbf{1.37$\times$} & \textbf{1.37$\times$} & \underline{1.02$\times$} \\
    \midrule

    \multicolumn{10}{l}{\textbf{Synthetic Task} \textit{(Datasets: PassageCount\_e, PassageRetrieval\_en\_e)}} \\
    \midrule
    Qwen2.5-7B-Instruct-1M & 15.00 & 99.33 & -- & 57.17 & 1.00$\times$ & 1.00$\times$ & 1.00$\times$ & 1.00$\times$ & 1.00$\times$ \\
    + LLMLingua2 & 6.33 & 99.00 & -- & 52.67 & 1.36$\times$ & 0.85$\times$ & 0.74$\times$ & 0.74$\times$ & 0.92$\times$ \\
    + SnapKV (pr.) & 16.60 & 97.67 & -- & 57.14 & 1.31$\times$ & \underline{1.55$\times$} & \underline{1.10$\times$} & \underline{1.10$\times$} & \underline{1.01$\times$} \\
    + D\textsuperscript{3} & 15.00 & 99.33 & -- & \underline{57.17} & 1.16$\times$ & 1.32$\times$ & 0.99$\times$ & 0.99$\times$ & 1.00$\times$ \\
    \rowcolor{gray!10}
    + \textbf{DASH (Ours)} & 16.67 & 98.00 & -- & \textbf{57.34} & \textbf{1.74$\times$} & \textbf{1.87$\times$} & \textbf{1.41$\times$} & \textbf{1.41$\times$} & \textbf{1.02$\times$} \\
    \midrule

    \multicolumn{10}{l}{\textbf{Code Completion} \textit{(Datasets: lcc\_e, RepoBench-P\_e)}} \\
    \midrule
    Qwen2.5-7B-Instruct-1M & 59.86 & 54.52 & -- & 57.19 & 1.00$\times$ & 1.00$\times$ & 1.00$\times$ & 1.00$\times$ & 1.00$\times$ \\
    + LLMLingua2 & 49.64 & 53.05 & -- & \underline{51.35} & \underline{1.53$\times$} & 0.67$\times$ & 0.84$\times$ & 0.84$\times$ & 0.97$\times$ \\
    + SnapKV (pr.) & 51.89 & 49.37 & -- & 50.63 & 1.39$\times$ & 1.15$\times$ & \underline{1.19$\times$} & \underline{1.19$\times$} & \underline{1.04$\times$} \\
    + D\textsuperscript{3} & 45.54 & 40.34 & -- & 42.94 & 1.20$\times$ & 1.04$\times$ & 1.09$\times$ & 1.09$\times$ & 1.00$\times$ \\
    \rowcolor{gray!10}
    + \textbf{DASH (Ours)} & 54.58 & 51.14 & -- & \textbf{52.86} & \textbf{1.86$\times$} & \textbf{1.25$\times$} & \textbf{1.40$\times$} & \textbf{1.46$\times$} & \textbf{1.04$\times$} \\

    \bottomrule
  \end{tabular}}
\end{table*}

\subsection{Vision--Language Benchmarks: Detailed Results and Efficiency Analysis}
\label{app:vl_details}

 As summarized in Table~\ref{tab:vit}, we compare DASH with representative advanced token compression methods on Qwen2-VL-7B across multiple general-purpose benchmarks under different compression ratios.
 
\textbf{Comparison between Different Methods:} 
Across a wide range of compression ratios and general-purpose benchmarks, \textbf{DASH consistently outperforms existing advanced token compression methods}, demonstrating its superior robustness under aggressive token reduction.
At a moderate compression ratio of 75.00\%, DASH achieves the highest average decline ratio (ADR) of 85.2, surpassing VisionZip, DART, and PruMerge+ by clear margins.
Even under more aggressive settings, such as 88.89\% compression, DASH maintains a strong overall performance (ADR of 74.7), while other methods exhibit noticeable degradation across multiple benchmarks.
These results indicate that DASH is more effective at identifying and preserving task-relevant visual tokens, enabling it to better balance efficiency and accuracy compared to prior approaches.

\textbf{Comparison between Different Compression Ratios:} 
As the compression ratio increases, all token compression methods inevitably experience performance degradation; however, \textbf{DASH demonstrates significantly more graceful degradation}.
At 93.75\% compression, DASH still achieves an ADR of 66.7, substantially outperforming FastV, VisionZip, and PruMerge+, whose ADR values drop below 60.
This trend highlights that DASH remains stable even when the majority of visual tokens are removed, suggesting that its compression strategy successfully retains the most informative visual content under extreme constraints.
The results further suggest that naive or heuristic pruning strategies struggle to scale to ultra-high compression regimes, whereas DASH maintains robustness across all evaluated settings.

\textbf{Comparison between Different Tasks:} 
When analyzing task-level performance, we observe that the advantages of DASH are particularly pronounced on general-purpose benchmarks such as GQA, MME, and MMStar, where global visual information is often sufficient for correct reasoning.
However, on tasks requiring fine-grained visual understanding—most notably OCRBench and ChartQA—the performance gap between DASH and other advanced methods narrows.
This observation is highly informative: while DASH effectively preserves globally salient visual cues, tasks involving dense text, precise numerical values, or structured visual layouts place higher demands on fine-grained local information.
Nonetheless, even under these challenging conditions, DASH remains competitive, indicating that its token selection mechanism captures both global and selectively important local features.

Overall, the comparisons across methods, compression ratios, and task types provide compelling evidence that \textbf{DASH achieves a more favorable accuracy--efficiency trade-off than existing token compression approaches}.
The consistent performance gains of DASH stem not from aggressive preservation of visual detail, but from its ability to adaptively retain the most informative tokens for a given input.
These findings also suggest that current general-purpose benchmarks predominantly reward global visual understanding, while more fine-grained tasks reveal the complementary strengths of sophisticated token compression strategies.

We evaluate inference efficiency on MMBench-EN by reporting total GPU latency and prefilling latency, together with FLOPs and KV cache usage (Table~\ref{tab:efficiency}). 
Our method achieves a favorable balance between accuracy and efficiency under both pruning settings. 
When retaining 35\% tokens in both the ViT and LLM stages, our latency drops to 60.8\% of the vanilla baseline (prefill latency 54.8\%), while maintaining competitive accuracy (77.5) compared with representative LLM-stage methods such as FastV, V$^2$Drop, and DART.
Under the more aggressive setting (ViT retain 20\% / LLM retain 20\%), our approach reduces latency to 51.7\% (prefill 52.1\%), with an accuracy of 73.9, showing a substantially improved speed--accuracy trade-off over ViT-stage pruning baselines (e.g., DART(ViT), ToMe, and ToFu) which achieve lower accuracy despite comparable or higher latency.

We also observe that prefilling latency generally decreases in tandem with total latency, indicating that the proposed pruning strategy effectively reduces the dominant computation in the prefilling stage.
While ViT-stage methods tend to achieve larger reductions in FLOPs, they often incur sharper accuracy degradation, whereas LLM-stage methods provide limited additional speedup once the runtime approaches a plateau.
Moreover, our method retains more LLM tokens than pure ViT-stage pruning approaches and thus exhibits a relatively larger KV cache footprint; nevertheless, this overhead remains moderate and practical for deployment.
Finally, the results suggest that FLOPs alone do not reliably predict end-to-end GPU latency, as methods with similar FLOPs can lead to noticeably different inference times due to kernel efficiency and memory behavior.

\begin{table*}[!t]
\small
\renewcommand{\arraystretch}{0.9}
\centering
\caption{Comparison of Advanced Token Compression Methods on Qwen2-VL-7B. ADR refers to the average decline ratio, which is the average value of the decline ratio of each benchmark (computed over GQA, MME, POPE, MMStar, OCRBench, ChartQA).}
\label{tab:vl_full}
\renewcommand{\tabcolsep}{8pt}
\resizebox{\textwidth}{!}{
\begin{tabular}{l|c c c c c c |c}
\toprule
\textbf{Method} & \textbf{GQA} & \textbf{MME} & \textbf{POPE} & \textbf{MMStar} & \textbf{OCRBench} & \textbf{ChartQA} & \textbf{ADR}\\
\midrule
Qwen2-VL-7B & \multicolumn{7}{c}{\textit{Upper Bound. All Tokens (100\%)}} \\ \midrule
Vanilla & 62.3 & 2306 & 88.4 & 57.1 & 80.7 & 81.6 & 100.0\\
\midrule

Qwen2-VL-7B & \multicolumn{7}{c}{\textit{Token Reduction ($\downarrow$ 75.00\%)}} \\ \midrule
+ FastV      & 57.3 & \underline{2101} & 83.7 & 44.8 & 41.8 & 58.9 & 80.0\\
+ VisionZip  & 58.6 & 2058 & \underline{86.8} & 47.1 & 42.5 & \textbf{66.2} & \underline{83.0}\\
+ PruMerge+  & \underline{59.1} & 2044 & \textbf{87.2} & \underline{48.1} & 34.2 & 55.8 & 79.5\\
+ DART       & 57.1 & 2056 & 84.8 & 46.9 & \textbf{52.0} & 53.1 & 81.4\\
\rowcolor{gray!10}
+ \textbf{DASH (Ours)} & \textbf{59.8} & \textbf{2109} & 83.5 & \textbf{55.4} & \underline{44.6} & \underline{62.6} & \textbf{85.2}\\
\midrule

Qwen2-VL-7B & \multicolumn{7}{c}{\textit{Token Reduction ($\downarrow$ 88.89\%)}} \\ \midrule
+ FastV      & 52.1 & 1866 & 77.5 & \underline{40.2} & 26.1 & 33.2 & 65.9\\
+ VisionZip  & 53.3 & 1819 & \underline{83.3} & 40.3 & 24.7 & \underline{47.9} & \underline{69.8}\\
+ PruMerge+  & \textbf{54.5} & 1794 & \textbf{84.1} & 38.5 & 22.4 & 44.5 & 68.4\\
+ DART       & 52.0 & \underline{1903} & 80.2 & 39.4 & \underline{41.2} & 31.0 & 69.1\\
\rowcolor{gray!10}
+ \textbf{DASH (Ours)} & \underline{53.9} & \textbf{1928} & 78.3 & \textbf{43.4} & \textbf{41.3} & \textbf{50.7} & \textbf{74.7}\\
\midrule

Qwen2-VL-7B & \multicolumn{7}{c}{\textit{Token Reduction ($\downarrow$ 93.75\%)}} \\ \midrule
+ FastV      & \underline{49.2} & 1693 & 75.0 & \underline{37.4} & 18.9 & 21.1 & 58.7\\
+ VisionZip  & 49.0 & 1704 & \textbf{80.1} & 35.3 & 15.7 & \underline{27.5} & 59.7\\
+ PruMerge+  & 48.7 & 1668 & \underline{79.2} & 33.4 & 14.3 & \textbf{30.2} & 58.9\\
+ DART       & 49.0 & \underline{1774} & 78.2 & 33.2 & \textbf{34.0} & 19.8 & \underline{61.4}\\
\rowcolor{gray!10}
+ \textbf{DASH (Ours)} & \textbf{53.9} & \textbf{1901} & 78.3 & \textbf{40.4} & \underline{32.5} & 25.7 & \textbf{66.7}\\

\bottomrule
\end{tabular}}
\end{table*}

\begin{table*}[!t]
  \centering
  \caption{Comparison of methods on Qwen2-VL. 
  \textbf{Avg. Acc.} denotes the average percentage of performance relative to Vanilla.}
  \resizebox{0.9\textwidth}{!}{
  \begin{tabular}{lcccccccc}
  \toprule
  \textbf{Methods} 
    & \textbf{GQA} 
    & \textbf{MME} 
    & \textbf{POPE} 
    & \textbf{SEED} 
    & \textbf{VQA\textsuperscript{text}} 
    & \textbf{OCRBench} 
    & \textbf{Avg. Acc.} \\
  \midrule
  Vanilla 
    & 61.5 & 2319 & 89.0 & 76.6 & 82.1 & 80.3 
    & 100\% \\
  
  \midrule
  \multicolumn{7}{c}{\textit{ViT retain 50\% tokens / LLM retain 50\% tokens}} \\
  \midrule
  
+ DART (ViT)   
    & 56.8 & \underline{2113} & 84.2 & 68.7 & \underline{63.6} & 46.9 
    & 87.2\% \\
+ ToMe          
    & 58.9 & 2007 & \textbf{88.2} & \underline{72.8} & 61.1 & 39.1 
    & 86.2\% \\
+ ToFu  
    & \textbf{59.1} & 1933 & \underline{88.1} & \textbf{73.2} & 60.6 & \textbf{76.7} 
    & 92.0\% \\
  \rowcolor{gray!10}
+ \textbf{DASH (Ours)} 
    & \textbf{59.1} & \textbf{2116} & 81.9 & 70.9 & \textbf{74.7} & \underline{60.0} 
    & \textbf{97.8\%} \\
  
  \midrule
  \multicolumn{7}{c}{\textit{ViT retain 35\% tokens / LLM retain 35\% tokens}} \\
  \midrule
  
+ DART (ViT)   
    & 55.2 & \underline{1968} & 81.1 & 64.3 & \underline{59.0} & \textbf{41.1} 
    & 82.3\% \\
+ ToMe          
    & \underline{58.3} & 1842 & \underline{87.4} & \underline{71.5} & 55.5 & 30.4 
    & 82.3\% \\
+ ToFu  
    & \textbf{58.6} & 1839 & \textbf{87.7} & \textbf{71.7} & 56.4 & 31.4 
    & 82.9\% \\
  \rowcolor{gray!10}
+ \textbf{DASH (Ours)} 
    & 55.4 & \textbf{1987} & 81.7 & 68.2 & \textbf{59.3} & \underline{39.1} 
    & \textbf{94.9\%} \\
  
  \bottomrule
  \end{tabular}
  }
  \label{tab:vit}
\end{table*}

\begin{table*}[!ht]
    \centering
    \caption{Inference efficiency comparison on MMBench-EN. Latency columns show relative percentage (normalized to Vanilla=100\%).}
    \resizebox{0.8\linewidth}{!}{
    \scriptsize
    \setlength{\tabcolsep}{1pt}
    \begin{tabular}{lccccc}
    \toprule
    \multirow{2}{*}{\textbf{Methods}} & \textbf{Latency}$\downarrow$ & \textbf{Prefill Latency}$\downarrow$ & \textbf{FLOPs}$\downarrow$ & \textbf{KV Cache}$\downarrow$ & \textbf{Accuracy}$\uparrow$ \\
    &(\%)&(\%)&(TFLOPs)&(MB)&(MMB)\\
    \midrule

    Vanilla & 100.0\% & 100.0\% & 31.9 & 76.6 & 80.5 \\

    \midrule
    \multicolumn{6}{c}{\textit{ViT retain 35\% tokens / LLM retain 35\% tokens}} \\
    \midrule
    \addlinespace

    + DART (ViT) & 67.1\% & 58.0\% & 11.7 & 29.7 & 70.6 \\
    + ToMe       & 65.9\% & 54.2\% & 12.0 & 30.4 & 72.6 \\
    + ToFu       & 67.5\% & 54.1\% & 12.3 & 29.6 & 73.9 \\
    + FastV      & 82.1\% & 74.3\% & 19.5 & 32.6 & 80.0 \\
    + V$^2$Drop  & 81.3\% & 74.6\% & 21.9 & 39.0 & 79.9 \\
    + SparseVLM  & 106.6\%& 94.8\% & 18.8 & 29.8 & 80.4 \\
    + DART       & 82.4\% & 78.6\% & 20.6 & 34.5 & 79.7 \\
    \rowcolor{gray!10}
    + \textbf{DASH (Ours)} & 70.6\% &74.8\% & 13.1 & 35.6 & 77.5 \\

    \midrule
    \multicolumn{6}{c}{\textit{ViT retain 20\% tokens / LLM retain 20\% tokens}} \\
    \midrule
    \addlinespace

    + DART (ViT) & 53.0\% & 53.8\% & 7.6  & 19.0 & 63.9 \\
    + ToMe       & 60.9\% & 53.2\% & 8.2  & 19.6 & 66.8 \\
    + ToFu       & 57.3\% & 53.0\% & 8.4  & 19.1 & 67.7 \\
    + FastV      & 76.6\% & 65.3\% & 17.0 & 23.0 & 77.4 \\
    + V$^2$Drop  & 74.2\% & 68.0\% & 18.8 & 27.6 & 78.1 \\
    + SparseVLM  & 92.4\% & 80.7\% & 16.4 & 20.6 & 79.2 \\
    + DART       & 75.0\% & 74.7\% & 18.0 & 25.0 & 77.8 \\
    \rowcolor{gray!10}
    + \textbf{DASH (Ours)} & 61.6\% & 62.5\% & 12.6 & 20.2 & 73.9 \\

    \bottomrule
    \end{tabular}
    }
    \label{tab:efficiency}
\end{table*}

\section{Additional Ablations and Extended Comparisons}
\label{sec:appendix_d}
\subsection{Start-Layer Selection and Proxy Validation}
\label{app:start_layer}

Treating the start layer $l_s$ as a hyperparameter raises the practical question of how to select it without exhaustive downstream sweeps. We use a lightweight LM proxy (perplexity-derived score) to rank candidate start layers. Within each proxy row, the \textbf{Top-1} layer is marked in \textbf{bold}, the \underline{Top-2} is underlined, and the \textit{Top-3} is italicized; within each downstream benchmark row, only the best-performing layer is bolded. A ``Top-$k$ hit'' occurs when the downstream-optimal layer is contained in the proxy's Top-$k$ shortlist, quantifying whether the proxy reliably narrows the search to a small constant-sized set.

Table~\ref{tab:proxy_hit} shows the hit-rates. At 20\% and 40\% token reduction, the Top-2/Top-3 shortlists consistently capture the downstream-optimal layer, supporting a two-stage protocol: (i) run the lightweight proxy to shortlist 2--3 candidates, and (ii) validate only those candidates on the downstream task. In addition, our empirical experience indicates that the best start layer consistently lies in $[0.3L, 0.5L]$, with $0.4L$ being best or near-best in most cases.

\begin{table*}[ht]
\small
\centering
\caption{Top-$k$ hit-rate of the perplexity proxy against the downstream-optimal start layer.}
\label{tab:proxy_hit}
\begin{tabular}{lccc}
\toprule
\textbf{Benchmark} & \textbf{Top-1 hit} & \textbf{Top-2 hit} & \textbf{Top-3 hit} \\
\midrule
LongBench-E (40\%) & 0 & 1 & 1 \\
LooGLE (40\%) & 1 & 1 & 1 \\
LongBench-E (20\%) & 1 & 1 & 1 \\
LooGLE (20\%) & 0 & 1 & 1 \\
\bottomrule
\end{tabular}
\end{table*}

Tables~\ref{tab:layer_sweep_20} and \ref{tab:layer_sweep_40} report the full layer sweeps at 20\% and 40\% reduction on Qwen2.5-7B-Instruct-1M. The proxy scores are perplexity-derived, with \emph{higher} values indicating better screening power in these measurements.

\begin{table*}[ht]
\small
\centering
\caption{Layer sweep at 20\% token reduction.}
\label{tab:layer_sweep_20}
\begin{tabular}{lccccc}
\toprule
\textbf{Metric} & \textbf{0.2$L$} & \textbf{0.3$L$} & \textbf{0.4$L$} & \textbf{0.5$L$} & \textbf{0.6$L$} \\
\midrule
LongBench-E avg. & 34.65 & \underline{43.27} & \textbf{46.25} & \textit{42.33} & 30.11 \\
ppl proxy (LB) & 20.32 & \textit{26.32} & \underline{28.31} & \textbf{28.64} & 21.73 \\
\midrule
LooGLE avg. & 10.64 & 16.32 & \textbf{19.94} & 17.43 & 12.23 \\
ppl proxy (LooGLE) & 9.56 & 14.30 & \textbf{15.22} & \underline{11.32} & 10.32 \\
\bottomrule
\end{tabular}
\end{table*}

\begin{table*}[ht]
\small
\centering
\caption{Layer sweep at 40\% token reduction.}
\label{tab:layer_sweep_40}
\begin{tabular}{lccccc}
\toprule
\textbf{Metric} & \textbf{0.2$L$} & \textbf{0.3$L$} & \textbf{0.4$L$} & \textbf{0.5$L$} & \textbf{0.6$L$} \\
\midrule
LongBench-E avg. & 30.10 & \underline{39.22} & \textbf{45.00} & \textit{32.44} & 10.23 \\
ppl proxy (LB) & 17.68 & \textit{22.12} & \textbf{26.39} & \underline{25.11} & 16.73 \\
\midrule
LooGLE avg. & \textit{14.43} & 13.32 & \textbf{18.24} & \underline{16.22} & 10.44 \\
ppl proxy (LooGLE) & 10.47 & \textbf{13.61} & \underline{13.56} & 9.37 & \textit{11.22} \\
\bottomrule
\end{tabular}
\end{table*}

\subsection{Kernel Compatibility and Ranking Fidelity}
\label{app:kernel_fidelity}

\paragraph{$\Delta_{\mathrm{attn}}$ vs. full-attention importance.} To isolate proxy fidelity from kernel advantages, we compared $\Delta_{\mathrm{attn}}$ ranking with a full-attention importance score (incoming attention mass, aggregated over heads) on Qwen2.5-7B over 500 long-context samples. Table~\ref{tab:attn_correlation} reports Spearman rank correlation and the IoU between halted token sets at 30\% sparsity. The agreement increases at deeper layers, providing strong evidence that $\Delta_{\mathrm{attn}}$ is a faithful proxy for full-attention-based ranking while remaining compatible with efficient kernels.

\begin{table*}[ht]
\small
\centering
\caption{Correlation between $\Delta_{\mathrm{attn}}$ ranking and full-attention importance ranking.}
\label{tab:attn_correlation}
\begin{tabular}{lcc}
\toprule
\textbf{Layer Depth} & \textbf{Spearman's $\rho$ ($\uparrow$)} & \textbf{Top-30\% Halting Overlap (IoU) ($\uparrow$)} \\
\midrule
Layer 10 (Early) & 0.74 & 82.4\% \\
Layer 16 (Middle) & 0.81 & 86.7\% \\
Layer 24 (Deep) & 0.88 & 91.2\% \\
\bottomrule
\end{tabular}
\end{table*}

\paragraph{Task-level ranking comparison.} Tables~\ref{tab:ranking_40} and \ref{tab:ranking_20} hold everything fixed (same model, data, start layer, retain ratio, seed, and Eager attention) and change only the ranking function. Under both 40\% and 20\% reduction, $\Delta_{\mathrm{attn}}$ ranking is competitive with (and here slightly better than) the full-attention importance-score baseline.

\begin{table*}[ht]
\scriptsize
\centering
\caption{LongBench-E at 40\% token reduction---$\Delta_{\mathrm{attn}}$ ranking vs. importance-score ranking (Eager).}
\label{tab:ranking_40}
\setlength{\tabcolsep}{3pt}
\begin{tabular}{lcccccccccccccc}
\toprule
\textbf{Method} & \textbf{Avg.} & \textbf{qasper} & \textbf{mfqa} & \textbf{hotpot} & \textbf{2wikimqa} & \textbf{gov.} & \textbf{mnews} & \textbf{trec} & \textbf{trivia} & \textbf{samsum} & \textbf{PC} & \textbf{PR} & \textbf{lcc} & \textbf{RepoBench} \\
\midrule
Vanilla & 48.87 & 44.19 & 51.13 & 62.97 & 51.07 & 6.97 & 6.57 & 65.00 & 85.75 & 32.95 & 15.00 & 99.33 & 59.86 & 54.52 \\
\midrule
\rowcolor{gray!10}
+ \textbf{$\Delta_{\mathrm{attn}}$ ranking} & 46.78 & 40.88 & 44.77 & 60.37 & 48.66 & 6.49 & 6.52 & 61.27 & 84.37 & 30.36 & 14.67 & 98.10 & 58.72 & 53.02 \\
+ Importance-score ranking & 46.03 & 38.14 & 45.22 & 60.24 & 50.01 & 7.11 & 6.30 & 62.60 & 85.25 & 32.87 & 13.70 & 98.23 & 53.45 & 45.32 \\
\bottomrule
\end{tabular}
\end{table*}

\begin{table*}[ht]
\scriptsize
\centering
\caption{LongBench-E at 20\% token reduction---$\Delta_{\mathrm{attn}}$ ranking vs. importance-score ranking (Eager).}
\label{tab:ranking_20}
\setlength{\tabcolsep}{3pt}
\begin{tabular}{lcccccccccccccc}
\toprule
\textbf{Method} & \textbf{Avg.} & \textbf{qasper} & \textbf{mfqa} & \textbf{hotpot} & \textbf{2wikimqa} & \textbf{gov.} & \textbf{mnews} & \textbf{trec} & \textbf{trivia} & \textbf{samsum} & \textbf{PC} & \textbf{PR} & \textbf{lcc} & \textbf{RepoBench} \\
\midrule
Vanilla & 48.87 & 44.19 & 51.13 & 62.97 & 51.07 & 6.97 & 6.57 & 65.00 & 85.75 & 32.95 & 15.00 & 99.33 & 59.86 & 54.52 \\
\midrule
\rowcolor{gray!10}
+ \textbf{$\Delta_{\mathrm{attn}}$ ranking} & 47.17 & 42.65 & 49.36 & 62.00 & 50.11 & 7.09 & 6.50 & 60.30 & 84.21 & 33.01 & 16.67 & 94.33 & 54.49 & 52.55 \\
+ Importance-score ranking & 46.69 & 39.14 & 45.98 & 61.54 & 50.31 & 7.00 & 6.60 & 62.60 & 85.25 & 33.87 & 16.70 & 97.23 & 52.33 & 48.37 \\
\bottomrule
\end{tabular}
\end{table*}

\subsection{Diagnostic Ablations}
\label{app:diagnostic}

\paragraph{Single-shot vs. multi-shot selection.} Table~\ref{tab:single_multi} compares single-shot selection with multi-shot variants under comparable pruning budgets on Qwen2.5-7B. Multi-shot provides only modest gains over single-shot (best 3-shot: +0.38 on LongBench-E and +0.27 on LooGLE), suggesting that single-shot captures most of the benefit with lower decision overhead.

\begin{table*}[ht]
\small
\centering
\caption{Single-shot vs. multi-shot selection on Qwen2.5-7B-Instruct-1M.}
\label{tab:single_multi}
\begin{tabular}{lcc}
\toprule
\textbf{Method} & \textbf{LongBench-E (Avg)} & \textbf{LooGLE (Avg)} \\
\midrule
Vanilla & 48.87 & 22.69 \\
\midrule
\rowcolor{gray!10}
+ \textbf{1-shot (Ours)} & 46.76 & 19.94 \\
+ 2-shot & 46.92 & 19.88 \\
+ 3-shot & 47.14 & 20.21 \\
+ 4-shot & 46.87 & 20.03 \\
\bottomrule
\end{tabular}
\end{table*}

\subsection{Extended Evaluations on Additional Models and Baselines}
\label{app:extended_eval}

\paragraph{Additional text backbones.} Tables~\ref{tab:new_models_20} and \ref{tab:new_models_40} evaluate DASH alongside SlimInfer and LazyLLM on Llama-3.1-8B, Qwen-3-8B, and Qwen2.5-7B-Instruct-1M at 20\% and 40\% token reduction. DASH remains competitive across model families while maintaining FlashAttention compatibility.

\begin{table*}[ht]
    \small
    \centering
    \caption{LongBench-E accuracy at 20\% token reduction on additional text backbones.}
    \label{tab:new_models_20}
    \begin{tabular}{lccc}
    \toprule
    \textbf{Method} & \textbf{Llama-3.1-8B} & \textbf{Qwen-3-8B} & \textbf{Qwen-2.5-7B-Instruct-1M} \\
    \midrule
    Vanilla & 49.23 & 57.33 & 48.87 \\
    \midrule
    + SlimInfer & \textbf{47.03} & \textbf{56.12} & \underline{44.70} \\
    + LazyLLM & 43.11 & 52.12 & 41.37 \\
    \rowcolor{gray!10}
    \textbf{+ DASH (Ours)} & \underline{46.11} & \underline{54.25} & \textbf{46.25} \\
    \bottomrule
    \end{tabular}
    \end{table*}

    \begin{table*}[ht]
    \small
    \centering
    \caption{LongBench-E accuracy at 40\% token reduction on additional text backbones.}
    \label{tab:new_models_40}
    \begin{tabular}{lccc}
    \toprule
    \textbf{Method} & \textbf{Llama-3.1-8B} & \textbf{Qwen-3-8B} & \textbf{Qwen-2.5-7B-Instruct-1M} \\
    \midrule
    Vanilla & 49.23 & 57.33 & 48.87 \\
    \midrule
    + SlimInfer & \textbf{45.54} & \textbf{54.36} & \textbf{47.12} \\
    + LazyLLM & \underline{44.74} & 51.23 & \underline{44.94} \\
    \rowcolor{gray!10}
    \textbf{+ DASH (Ours)} & 43.12 & \underline{52.23} & 44.00 \\
    \bottomrule
    \end{tabular}
  \end{table*}

\paragraph{Stronger VL backbones.} Tables~\ref{tab:qwen25vl_25} and \ref{tab:internvl_25} report results on Qwen2.5-VL-7B and InternVL-2.5-8B under a 25\% prefill pruning budget. DASH stays competitive on both strong MLLMs.

\begin{table*}[ht]
\small
\centering
\caption{Qwen2.5-VL-7B at 25\% token reduction.}
\label{tab:qwen25vl_25}
\begin{tabular}{lcccccc}
\toprule
\textbf{Method} & \textbf{MME} & \textbf{POPE} & \textbf{Nocaps} & \textbf{OKVQA} & \textbf{VizWiz} & \textbf{VQAv2} \\
\midrule
Vanilla (100\%) & 2349 & 90.49 & 1.16 & 64.56 & 63.51 & 81.70 \\
\midrule
+ FastV (25\%) & 2251 & 89.76 & 1.14 & 63.89 & 62.91 & 79.46 \\
+ DART (25\%) & 2252 & 89.70 & 1.11 & 63.83 & 62.66 & 79.47 \\
+ DivPrune (25\%) & 2189 & 89.91 & 1.11 & 63.46 & 62.29 & 79.23 \\
\rowcolor{gray!10}
+ \textbf{DASH (Ours, 25\%)} & 2287 & 88.93 & 1.13 & 61.73 & 61.46 & 80.58 \\
\bottomrule
\end{tabular}
\end{table*}

\begin{table*}[ht]
  \small
  \centering
  \caption{InternVL-2.5-8B at 25\% token reduction.}
  \label{tab:internvl_25}
  \begin{tabular}{lccccccc}
  \toprule
  \textbf{Method} & \textbf{AI2D} & \textbf{ChartQA} & \textbf{DocVQA} & \textbf{MME} & \textbf{MMStar} & \textbf{POPE} & \textbf{TextVQA} \\
  \midrule
  Vanilla (100\%) & 74.10 & 77.40 & 91.00 & 2240 & 58.40 & 87.70 & 76.23 \\
  \midrule
  + FastV (25\%) & 67.49 & \underline{63.68} & 71.26 & 2073 & 49.26 & 81.54 & \textbf{74.65} \\
  + DART (25\%) & 62.37 & 49.44 & 52.99 & 2086 & 50.58 & 82.53 & 64.74 \\
  + DivPrune (25\%) & \underline{69.95} & 58.84 & \textbf{73.88} & \underline{2097} & \underline{52.56} & \underline{83.98} & 70.43 \\
  \rowcolor{gray!10}
  + \textbf{DASH (Ours, 25\%)} & \textbf{70.27} & \textbf{65.34} & \underline{72.18} & \textbf{2098} & \textbf{53.58} & \textbf{84.03} & \underline{73.28} \\
  \bottomrule
  \end{tabular}
  \end{table*}

\paragraph{Dynamic-LLaVA comparison.} Table~\ref{tab:dynamic_llava} provides a direct comparison with Dynamic-LLaVA-7B at 25\% reduction. Dynamic-LLaVA is a training-based approach, whereas DASH is training-free and applies at inference time without extra optimization.

\begin{table*}[ht]
  \small
  \centering
  \caption{Dynamic-LLaVA-7B at 25\% token reduction.}
  \label{tab:dynamic_llava}
  \begin{tabular}{lcccccc}
  \toprule
  \textbf{Method} & \textbf{VQAv2} & \textbf{GQA} & \textbf{SciQA} & \textbf{TextVQA} & \textbf{POPE} & \textbf{MMBench} \\
  \midrule
  Vanilla & 78.5 & 62.0 & 66.8 & 58.2 & 85.9 & 64.3 \\
  \midrule
  + ToMe & 66.0 & --- & 62.7 & 56.0 & 51.0 & 56.9 \\
  + ATS & 66.7 & --- & 63.0 & 55.1 & 57.4 & 54.9 \\
  + EViT & 65.5 & --- & 64.1 & 54.2 & 60.1 & 56.2 \\
  + LLaVA-PruMerge+ & 76.8 & --- & 68.3 & \underline{57.1} & \underline{84.0} & \underline{64.9} \\
  + LLaVA-FastV & 75.1 & 57.5 & \underline{68.7} & 56.2 & 81.0 & 63.5 \\
  + VoCo-LLaMA & \underline{76.9} & 59.8 & --- & --- & --- & 61.0 \\
  + LLaVA-HiRED & 74.7 & 59.8 & 66.4 & 44.2 & --- & --- \\
  + Dynamic-LLaVA-7B\_I & \textbf{78.0} & \textbf{61.4} & \textbf{69.1} & 57.0 & \textbf{85.0} & \textbf{65.4} \\
  \rowcolor{gray!10}
  + \textbf{DASH (Ours)} & 75.6 & \underline{60.3} & 64.9 & \textbf{57.6} & 82.5 & 61.8 \\
  \bottomrule
  \end{tabular}
  \end{table*}

\paragraph{Eager-only baseline comparison.} To remove FlashAttention-related kernel confounds, we reran the key LongBench-E comparison under a fully Eager configuration. Table~\ref{tab:eager_baseline_avg} reports the average scores, and Table~\ref{tab:eager_baseline_full} provides the per-task breakdown.

\begin{table*}[ht]
\small
\centering
\caption{Eager-only baseline comparison on LongBench-E (average scores).}
\label{tab:eager_baseline_avg}
\begin{tabular}{lc}
\toprule
\textbf{Method} & \textbf{LongBench-E Avg} \\
\midrule
Vanilla & 48.87 \\
\midrule
+ SnapKV(pr.) & 46.24 \\
+ FastV & 43.93 \\
\rowcolor{gray!10}
+ \textbf{DASH (Ours)} & 46.78 \\
\bottomrule
\end{tabular}
\end{table*}

\begin{table*}[ht]
  \scriptsize
  \centering
  \caption{Eager-only baseline comparison on LongBench-E (per-task breakdown).}
  \label{tab:eager_baseline_full}
  \setlength{\tabcolsep}{3pt}
  \begin{tabular}{lcccccccccccccc}
  \toprule
  \textbf{Method} & \textbf{Avg.} & \textbf{qasper} & \textbf{mfqa} & \textbf{hotpot} & \textbf{2wikimqa} & \textbf{gov.} & \textbf{mnews} & \textbf{trec} &
  \textbf{trivia} & \textbf{samsum} & \textbf{PC} & \textbf{PR} & \textbf{lcc} & \textbf{RepoBench} \\
  \midrule
  Vanilla & 48.87 & 44.19 & 51.13 & 62.97 & 51.07 & 6.97 & 6.57 & 65.00 & 85.75 & 32.95 & 15.00 & 99.33 & 59.86 & 54.52 \\
  \midrule
  + SnapKV(pr.) & \underline{46.24} & 38.88 & 42.45 & \textbf{61.56} & \underline{48.34} & \textbf{7.00} & \textbf{6.78} & \textbf{63.67} & \textbf{85.64} &
  \underline{31.01} & \textbf{16.12} & \underline{97.62} & \underline{52.89} & 49.17 \\
  + FastV & 43.93 & \textbf{41.02} & \underline{42.63} & 55.22 & 41.42 & \underline{6.86} & 6.21 & \underline{61.33} & 83.34 & \textbf{33.42} & 9.63 & 85.02
  & 52.71 & \underline{52.31} \\
  \rowcolor{gray!10}
  + \textbf{DASH (Ours)} & \textbf{46.78} & \underline{40.88} & \textbf{44.77} & \underline{60.37} & \textbf{48.66} & 6.49 & \underline{6.52} & 61.27 & \underline{84.37} & 30.36 & \underline{14.67} & \textbf{98.10} & \textbf{58.72} & \textbf{53.02} \\
  \bottomrule
  \end{tabular}
  \end{table*}

\end{document}